\documentclass[sigconf]{acmart}

\AtBeginDocument{%
  \providecommand\BibTeX{{%
    \normalfont B\kern-0.5em{\scshape i\kern-0.25em b}\kern-0.8em\TeX}}}

\setcopyright{acmlicensed}
\copyrightyear{2024}
\acmYear{2024}
\acmDOI{X}

\acmConference[KDD '24]{ACM Knowledge Discovery and Data Mining Conference}{August 25--29,
  224}{Barcelona, Spain}
\acmISBN{978-1-4503-XXXX-X/18/06}

\usepackage{amsmath}
\mathchardef\mhyphen="2D %
\usepackage{enumitem}
\usepackage{algorithm2e}
\RestyleAlgo{ruled}
\usepackage{mathtools}
\DeclarePairedDelimiter\floor{\lfloor}{\rfloor}

\usepackage{graphicx}
\usepackage{subcaption}
\usepackage{multirow}
\usepackage{colortbl} 

\usepackage{xcolor,color,soul,bm}

\newcommand{\rev}[1]{\textcolor[rgb]{0.00,0.00,0.00}{#1}}

\begin{document}

\title[Using Self-supervised Learning can Improve Model Fairness]{Using Self-supervised Learning Can Improve Model Fairness}

\author{Sofia Yfantidou}
\authornote{Work done at Nokia Bell Labs.}
\email{syfantid@csd.auth.gr}
\orcid{0000-0002-5629-3493}
\affiliation{%
  \institution{Aristotle University of Thessaloniki}
  \country{Greece}
}

\author{Dimitris Spathis}
\authornote{Also affiliated with the University of Cambridge, UK.}
\email{dimitrios.spathis}
\email{@nokia-bell-labs.com}
\orcid{0000-0001-9761-951X}
\affiliation{%
  \institution{Nokia Bell Labs}
  \country{United Kingdom}
}

\author{Marios Constantinides}
\authornotemark[2]
\email{marios.constantinides}
\email{@nokia-bell-labs.com}
\orcid{0000-0003-1454-0641}
\affiliation{%
  \institution{Nokia Bell Labs}
  \country{United Kingdom}
}

\author{Athena Vakali}
\email{avakali@csd.auth.gr}
\orcid{0000-0002-0666-6984}
\affiliation{%
  \institution{Aristotle University of Thessaloniki}
  \country{Greece}
}

\author{Daniele Quercia}
\email{daniele.quercia}
\email{@nokia-bell-labs.com}
\orcid{0000-0001-9461-5804}
\affiliation{%
  \institution{Nokia Bell Labs}
  \country{United Kingdom}
}

\author{Fahim Kawsar}
\email{fahim.kawsar}
\email{@nokia-bell-labs.com}
\orcid{0000-0001-5057-9557}
\affiliation{%
  \institution{Nokia Bell Labs}
    \country{United Kingdom}
}

\renewcommand{\shortauthors}{Yfantidou, et al.}
\urlstyle{tt}

\begin{abstract}
Self-supervised learning (SSL) has become the de facto training paradigm of large models, where pre-training is followed by supervised fine-tuning using domain-specific data and labels.
Despite demonstrating comparable performance with supervised methods, comprehensive efforts to assess SSL's impact on machine learning fairness (i.e., performing equally on different demographic breakdowns) are lacking. 
Hypothesizing that SSL models would learn more generic, hence less biased representations, this study explores the impact of pre-training and fine-tuning strategies on fairness. We introduce a fairness assessment framework for SSL, comprising five stages: defining dataset requirements, pre-training, fine-tuning with gradual unfreezing, assessing representation similarity conditioned on demographics, and establishing domain-specific evaluation processes.
We evaluate our method's generalizability on three real-world human-centric datasets (i.e., MIMIC, MESA, and GLOBEM) by systematically comparing hundreds of SSL and fine-tuned models on various dimensions spanning from the intermediate representations to appropriate evaluation metrics. 
 Our findings demonstrate that SSL can significantly improve model fairness, while maintaining performance on par with supervised methods---exhibiting up to a 30\% increase in fairness with minimal loss in performance through self-supervision. We posit that such differences can be attributed to representation dissimilarities found between the best- and the worst-performing demographics across models---up to $\times13$ greater for protected attributes with larger performance discrepancies between segments. \\\texttt{\rev{Code:}} \rev{\url{https://github.com/Nokia-Bell-Labs/SSLfairness}}
\end{abstract}

\maketitle

\section{Introduction}
Self-supervised learning (SSL) has emerged as a dominant training paradigm for large models, involving unsupervised pre-training followed by supervised fine-tuning using domain-specific data and labels. SSL has proven its performance robustness and beyond state-of-the-art capabilities mainly in the areas of computer vision (CV) \cite{chen2020simple} and natural language processing (NLP) \cite{devlin2018bert} research. Inspired by such efforts, which leverage massive amounts of unlabeled data, many research communities that deal with human-centric data have swiftly recognized the potential of self-supervision. SSL exploitation is promising to explore extensive unlabeled data, effectively complementing small, labeled domain datasets \cite{tang2020exploring}. As a case in point, in the healthcare domain, leveraging this wealth of unlabeled information can uncover intricate physiological and behavioral patterns at an unprecedented scale, offering novel insights into personalized and proactive healthcare \cite{perez2021wearables}.

Due to the recency of SSL adoption for human-centric, multimodal data, such as time-series, performance metrics, such as accuracy scores, are typically used as the main evaluation criteria. Yet, a performance-centric evaluation approach can result in discriminatory impacts when comparing across different demographics. For instance, in the context of supervised learning, \citet{kamulegeya2019using} found that neural network algorithms trained to perform skin lesion classification showed approximately half the original diagnostic accuracy on black patients. At the same time, people of color are consistently misclassified by health sensors such as oximeters as they were validated on predominantly white populations~\cite{sjoding2020racial}. 

Preliminary evidence suggests that SSL models may avoid such pitfalls due to their pre-training without (potentially) biased human annotations \cite{ramapuram2021evaluating}. Similarly, self-supervision has demonstrated superiority in key aspects of Data-centric Machine Learning (ML), namely a subset of Responsible ML, emphasizing data quality, such as robustness and uncertainty estimation \cite{hendrycks2019using}.
Yet, comprehensive efforts to compare the fairness of supervised and SSL models are lacking. Note that \rev{group} fairness assessments typically look for accuracy disparities among diverse protected attributes, namely sensitive personal characteristics, such as race or gender, that are legally safeguarded from discrimination. 

\begin{figure*}[tbh!]
  \centering
  \includegraphics[page=4,trim={0 0.8in 1.4in 0},clip,width=.75\linewidth]{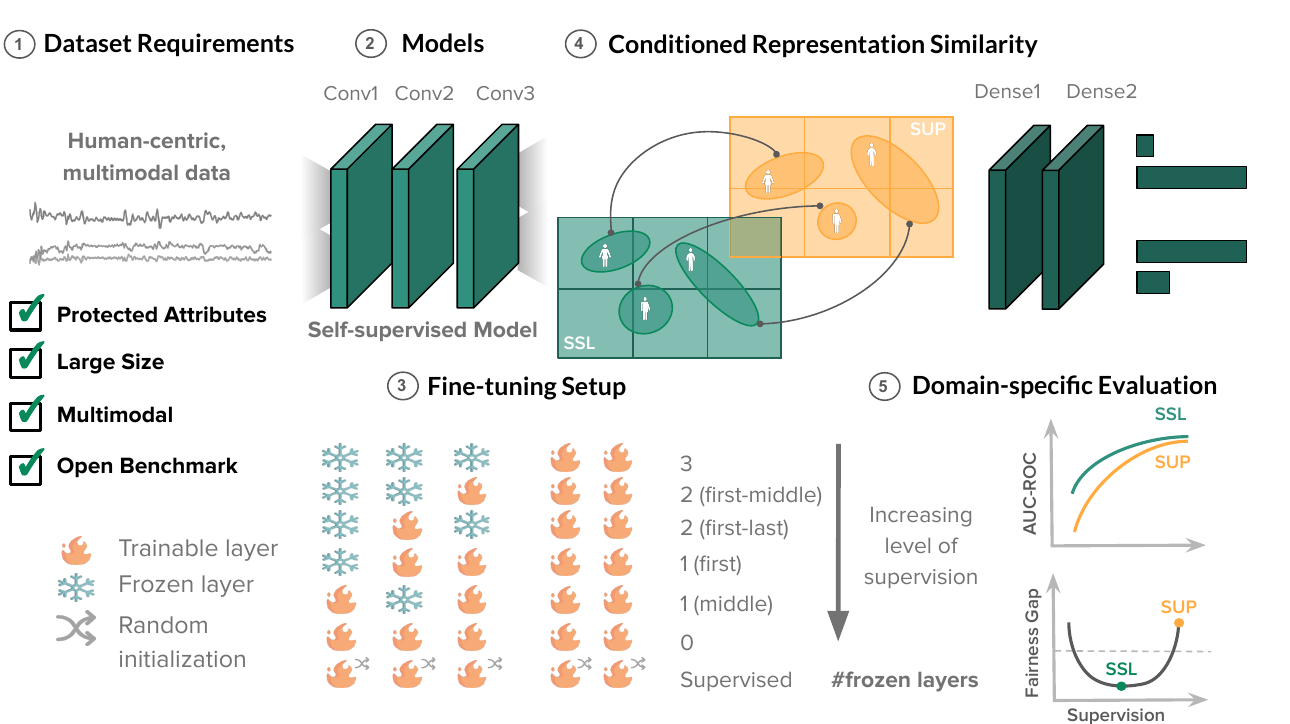}
  \caption{\textbf{Overview of the proposed fairness assessment framework for SSL.} Starting with benchmark  selection, we systematically study the impact of fine-tuning on fairness through a novel combination of evaluation and representation learning metrics.} 
  \label{fig:modelsummaries} 
  \vspace{-0.2cm}
\end{figure*}
In this work, we aim to bridge this gap by introducing a five-stage fairness assessment framework (Figure~\ref{fig:modelsummaries}) for SSL. \rev{Our framework encompasses} dataset requirements definition, modeling, fine-tuning setup, representation similarity, and domain-specific evaluation considerations. \rev{As such, it} facilitates an examination of how SSL fine-tuning affects fairness compared to supervised alternatives, focusing on both outcomes and representations. We hypothesize, that SSL models will exhibit less bias given that their representations are only partially affected by labels, which may comprise biases steered by the downstream tasks. \rev{Using SSL may seem less prone to bias, but concluding it is inherently fair oversimplifies. The pre-training phase can still encode biases from data distributions. Fine-tuning using labeled data can lead to bias amplification. The influence of contrastive objectives on fairness, versus the role of design choices like data augmentation, remains unexplored. These key aspects make the assessment of our hypothesis non-trivial.} In detail, we make four contributions:\footnote{\rev{Note that some results appeared as a workshop paper at the HCRL (AAAI 2024) \cite{yfantidou2024evaluating}}}
\vspace{-0.1cm}
\begin{enumerate}
    \item Moving away from conventional performance-centric assessments, we introduce a fairness assessment framework for SSL, integrating fairness metrics into our methodology to evaluate how fine-grained differences in the layer, model, and metric level between supervised and SSL models affect model outcomes and representations (\S\ref{sec:method}). 
    \item We conduct a systematic comparison of more than 100 models with various levels of supervision and fine-tuning on three large real-world benchmarks and tasks (\S\ref{sec:evaluation}). To foster reproducibility we make our code publicly available.\footnote{Code: \url{https://github.com/Nokia-Bell-Labs/SSLfairness}} 
    \item We show that SSL yields smaller performance discrepancies between groups, while performing on par with supervised models across datasets. More notably, we observe up to a 30\% increase in fairness, accompanied by only a 2\% loss in performance for certain SSL fine-tuning strategies. Similarly, the SSL model shows quicker fairness gains than the supervised one as a function of limited training data (\S\ref{sec:results:fairness}). 
    \item In light of these results, we compare learned representations using the latent similarity between supervised and SSL models, which reveals discrepancies in the latent space across different demographic groups. Specifically, the larger the performance gap between segments the larger the representation similarity gap (up to $\times13$ greater) between the SSL and the supervised models (\S\ref{sec:results:representation}).
\end{enumerate}
\section{Background \& Related Work}

\subsection{Bias \& Fairness in Machine Learning}\label{sec:fundamentals-bias}
There are two opposing perspectives when quantifying group fairness, i.e., statistical parity for individuals belonging to different protected groups \cite{dwork2012fairness}, in ML research: ``We're All Equal'' (WAE) and ``What You See Is What You Get'' (WYSIWYG) \cite{friedler2021possibility,10.1145/3442188.3445892}. The WAE perspective assumes equal ability across groups to perform a task and is closely related to treating equals equally. On the other hand, the WYSIWYG viewpoint assumes that the data itself reflects a group's ability with respect to the task, and thus, unequals should not be treated equally. Different fairness metrics quantify each perspective \cite{garg2020fairness}; demographic parity metrics, such as disparate impact and statistical parity difference, quantify WAE. Equality of odds metrics, such as average odds and average absolute odds difference, quantify WYSIWYG. However, the choice of metric is often guided by the question ``What is the consequence of the predictive outcome?.'' Equality of opportunity metrics, such as false negative rate, and false positive rate ratios, find common ground between the two perspectives. To capture the different perspectives, in this work, we utilize a combination of metrics, as discussed in \S\ref{sec:evaluation}.

\subsection{Fairness in Self-Supervised Learning}\label{sec:related-work}
While SSL methods (e.g., SimCLR \cite{chen2020simple}, BYOL \cite{grill2020bootstrap}, Masked Autoencoders \cite{he2022masked}) have seen widespread use in CV \cite{kolesnikov2019revisiting}, NLP \cite{lan2019albert}, and audio \cite{saeed2021contrastive},
they have also been validated in multimodal, human-centric data; yet, the area remains under-explored \cite{haresamudram2022assessing,spathis2021self}.

Existing works have extensively benchmarked SSL algorithms across domains, primarily focusing on performance metrics. However, limited attention has been given to evaluating fairness in SSL methods, particularly for multimodal, human-centric data. For example, in the healthcare setting, SSL has been applied to online patient monitoring \cite{yeche2021neighborhood}, Atrial Fibrillation detection \cite{tonekaboni2021unsupervised}, mortality or decompensation prediction \cite{harutyunyan2019multitask}, maternal and fetal stress detection \cite{sarkar2021detection}, and human-activity recognition \cite{tang2020exploring}, among others. Yet, the above works focus on performance-centric assessments.

However, the mere absence of biased annotations in SSL does not guarantee fairness, necessitating evaluations that extend beyond accuracy. Preliminary research efforts show that SSL techniques can incorporate protected attributes into their representations causing potentially unfair predictions on downstream tasks \cite{ma2021conditional}. For example, \citet{steed2021image} have demonstrated that image representations learned with unsupervised pre-training exhibit human-like biases. Yet, while studies in CV and audio have found similarities in intermediate representations between SSL and supervised alternatives, it is crucial to emphasize that most comparisons focus on aspects other than fairness \cite{grigg2021self,chung2019unsupervised}. To date, there is a distinct lack of comprehensive investigations specifically addressing fairness considerations in SSL learned representations.

Fairness evaluations in SSL have been more prevalent in CV \cite{ramapuram2021evaluating} and NLP, including recent advances in generative models \cite{sheng2019woman}. 
For instance, while models fine-tuned on top of pre-trained models can inherit their biases \cite{wang2023overwriting}, \citet{ranjit2023variation} have shown that supervised models tend to preserve their pre-training biases regardless of the target dataset, in contrast to SSL methods, where the fine-tuning objective and dataset influence the extent of transferred biases. Discussions on SSL's impact on fairness include considerations of training without prior data curation and the effects of fine-tuning \cite{goyal2022vision,ramapuram2021evaluating,mao2023last}.
However, in multimodal, human-centric data, fairness evaluations have seen limited exploration, mainly in a supervised setting. For instance, for in-hospital mortality using the MIMIC dataset \cite{meng2022interpretability,roosli2022peeking} or keyword spotting for on-device ML \cite{10.1145/3591867}.
Hence, such efforts are still in their early stages \cite{yfantidou2023beyond}. 

\noindent\textbf{Research Gap.} This paper aims to address the research gap by assessing SSL approaches in real-world human-centric data, considering both performance and fairness aspects.
While there are existing works addressing performance, fairness, or learned representations \textit{individually} in SSL (across different domains), evidence that connects all these three aspects, particularly in human-centric, multimodal, data, such as time-series, is still lacking.

\section{Method}\label{sec:method}
In this section, we introduce the proposed framework to investigate how design choices in SSL affect outputs and representations. 
\smallskip

\noindent\textbf{Notation.} Let $X=(x_1,\ldots ,x_N)\in \mathbb{R}^{N\times T\times M}$ denote  an input sequence with $N$ samples of $T$ sample length and $M$ modalities (e.g., multivariate signals), and $Y=(y_1,\ldots ,y_N)\in \mathbb{R}^N$ denote the respective binary output sequence.  
In the context of SSL, let $f(\cdot)$ denote an encoder that maps input samples $X$ into intermediate embeddings $H=(h_1,\ldots ,h_D)\in \mathbb{R}^{N\times D}$ where $D$ is the size of the latent dimension. These embeddings are further trained by the fine-tuning strategy $\phi$ employed for the downstream task resulting in $H_\phi=(h_{1,\phi},\ldots,h_{D,\phi})\in \mathbb{R}^{N\times D}$, where $\Phi=\{\phi_1$,\ldots,$\phi_L\}\in \mathbb{B}$ controls the training status (trainable or frozen) of each layer $l$ in the base encoder, where $L$ is the total number of layers. For conditioning these representations on protected attributes, we denote $P$ as the set of protected attributes, and $V_p$ as the set of possible values for the protected attribute $p$. Thus, we denote the conitioned representations as $H_{\phi,p=v}=(h_{1,\phi,p=v},\ldots , h_{D,\phi,p=v}) \in \mathbb{R}^{N'\times D}$, where $N'$ is the subgroup of samples, where the user has a value $v$ for the protected attribute $p$.

\rev{Put simply, SSL models learn representations from data in an unsupervised manner, potentially avoiding biases present in labeled data. In contrast, supervised models can amplify biases in labels into learned representations. SSL's contrastive learning objective encourages invariant representations, aligning with debiasing goals. The subsequent supervised fine-tuning has a limited capacity to bias precomputed representations compared to training a supervised model from scratch. We hypothesise that the information bottleneck theory \cite{shwartz2017opening} in SSL acts as a regularizer, potentially mitigating biased signals during pretraining.}
In the following, we present the five stages of our framework for facilitating fairness assessments in SSL:
\smallskip

\noindent\textbf{1. Dataset Requirements Definition.} In the context of fairness analyses and the scope of this work, it is essential to consider certain requirements during benchmark dataset selection, significantly limiting the available dataset choices as follows:
\begin{enumerate}
    \item \textbf{Protected attributes:} The dataset should provide at least one protected attribute, such as age, gender, or race;
    \item \textbf{Size:} The dataset should contain data from ``sufficient''\footnote{Considering that SSL requires large datasets for pre-training, we focus on large human-centric datasets with thousands of samples.} users to allow for statistical comparisons (of fairness and performance metrics) between user segments;
    \item \textbf{Modality:} The dataset should contain more than one modality, such as different sensor measurements, hence excluding unimodal benchmarks in vision or language;
    \item \textbf{Open Benchmark:} To foster reproducibility and allow for comparisons with the literature, we focus on publicly available benchmarks and pre-processing pipelines. %
\end{enumerate}

\noindent\textbf{2. Models.} 
We use a SimCLR \cite{chen2020simple} variant adapted for time-series data \cite{tang2020exploring}. Our design mirrors SimCLR's components: a) a \textit{stochastic data augmentation} module that transforms a data sample $x$ in two correlated views, denoted $\tilde{x}_i$ and $\tilde{x}_j$ (i.e., positive pair) by employing scaling and signal inversion; b) a \textit{base encoder} $f(\cdot)$ for extracting representation embeddings from augmented data samples. We opt for a 3-layer Convolutional Neural Network (CNN) in line with \cite{tang2020exploring} to obtain $h_i=f(\tilde{x}_i)=\text{ConvNet}(\tilde{x}_i)$, where $h_i\in \mathbb{R}^d$ is the output after the max pooling layer (dependent on the fine-tuning setup described in the next section); c) a \textit{projection head} $g(\cdot)$ for mapping representations to the contrastive loss space. We opt for a 2-layer Multi-layer Perceptron (MLP) to obtain $z_i^{[l]}=g(h_i^{[l]})=W^{[l]}\sigma(^{[l-1]}h_i^{[l-1]})$, where $\sigma$ is a ReLU non-linearity and $l=3$ for 2 hidden and 1 output layers; d) a contrastive loss function, namely a normalized temperature-scaled cross-entropy loss (NT-Xent) \cite{sohn2016improved,chen2020simple}. 
We define a similar architecture for the supervised baseline, replacing the contrastive loss with categorical cross-entropy.\hfill
\smallskip

\noindent\textbf{3. Fine-tuning Setup.} To assess the effect of self-supervision on fairness outcomes and representations, we employ a ``gradual unfreezing'' strategy (Algorithm~\ref{alg:one}) \cite{howard-ruder-2018-universal}, balancing the impact of the pre-trained encoder and the downstream labels. We start by freezing all three base encoder layers and fine-tuning only the projection head. Then, starting from the last layer containing the least general knowledge \cite{yosinski2014transferable}, we gradually unfreeze layers one by one (or block by block, achieved via a step of $\floor*{\frac{L}{3}}$ in the pseudocode) until the encoder layers and projection head are fully trainable (similar to full supervision). We also experiment with different freezing configurations, similar to ``surgical fine-tuning'' \cite{lee2022surgical}, where we tune only one (block of) layer(s), and freeze the remaining, as tuning different blocks of layers performs best for different types of distribution shifts. Figure~\ref{fig:modelsummaries} visualizes the described fine-tuning setup.
\SetKwComment{Comment}{/* }{ */}
\SetKw{KwBy}{by}
\begin{algorithm}
\caption{Gradual Unfreezing}\label{alg:one}
\KwIn{Sequence $X$ and encoder $f(\cdot)$ where the layers' training status is controlled by $\Phi=\{\phi_1$,\ldots,$\phi_L\}$} 
\KwOut{Embeddings $H_\phi=(h_{1,\phi},\ldots,h_{D,\phi})$} 
\For{$l\gets L-1$ \KwTo $0$}{
    $\Phi[l] \gets 0$\ \Comment*[r]{freeze all}
    }
\For{$l\gets L-1$ \KwTo $0$ \KwBy $\floor*{\frac{L}{3}}$}{
    $\Phi[l] \gets 1$\ \Comment*[r]{unfreeze one-by-one}
    $H_\phi \gets f(X,\Phi)$ \Comment*[r]{fine-tune trainable}
    }
\end{algorithm}

More formally, we define individual parameters $\Phi=\{\phi_1$, $\phi_2$, $\phi_3$\} to control the training status of each layer in the 3-layer base encoder. The output $h_i$ is then determined by the ConvNet function with parameters $\Phi$, where $\phi_i$ indicates whether each corresponding layer is frozen (0) or trainable (1). This allows for a flexible downstream task configuration where specific layers can be selectively frozen or trained based on the desired experimental setup, where the representations embeddings $H_\phi$ are obtained as follows:
$$
h_{i,\Phi} = f(\tilde{x}_i, \Phi) = \text{ConvNet}(\tilde{x}_i, \Phi)
$$
\noindent\textbf{4. Custom Representation Similarity Function.}
For assessing the impact of supervision on learned representations, we adopt the linear Centered Kernel Alignment (CKA) method as a similarity index. CKA has proven superior to related methods, such as linear regression, or canonical correlation analysis (CCA), addressing challenges regarding the distributed nature, potential misalignment, and high dimensionality of representations \cite{kornblith2019similarity}. We propose the conditioning of CKA on protected attributes to identify differences in representation similarity between diverse demographic groups.

More formally, let $P$ represent the set of protected attributes, and $V_p$ represent the set of possible values for the protected attribute $p$. Given the $H_\phi\in \mathbb{R}^{N\times D}$ activations for the SSL model (i.e., intermediate feature representations) and $J\in \mathbb{R}^{N\times D}$ activations for the supervised model, for the same examples, we can calculate CKA based on a subset of those activations conditioned on the users' protected attributes as $H_{\phi,p=v} = \{ h_{i,\Phi} \mid p(\tilde{x}_i) = v, \, p \in \mathcal{P}, \, v \in \mathcal{V}_p \}$. Then, the conditioned linear CKA is given by:
$$
CKA(H,J,P,V)=\frac{\lVert H_{\phi,p=v}^TJ_{p=v}\rVert_F^2}{\lVert H_{\phi,p=v}^TH_{\phi,p=v}\rVert_F\lVert J_{p=v}^TJ_{p=v}\rVert_F}
$$

\noindent\textbf{5. Domain-specific Evaluation Processes.}
Enhancing fairness in ML requires a means to quantify biases. Particularly in human-centric settings and high-stakes applications, single evaluation metrics struggle to reflect the success of ML models. As such, monitoring and reporting a multitude of metrics across different protected groups becomes the norm. Fairness trees \cite{saleiro2018aequitas} accompanied by domain expertise can help researchers choose appropriate metrics.

To capture the different fairness perspectives (\S\ref{sec:fundamentals-bias}), we adopt multiple ratio-based metrics. Specifically, we use the disparate impact (WAE), false omission rate, false discovery rate, false negative rate, and false positive rate (hybrid) ratios. We adopt the above metrics to acknowledge the potential consequences of both false positives and false negatives in the context of healthcare and the implications of prediction disparities among protected attributes (e.g., falsely administered medication or unnecessary financial burden for false positives and life threatening consequences or missed treatment opportunities for false negatives) \cite{burt2017burden,dresselhaus2002ethical}. 

Ratio metrics are bounded within the range $[0,+\infty)$, where a value of 1.0 signifies parity across protected attributes. In this work, we define and use a custom fairness meta-metric, the so called parity deviation, as a fairness indicator, which we calculate as follows:
$$\textit{Parity Deviation}_{fairness\_metric}=\|1-fairness\_metric\|$$
Here, the term ``metric'' refers to the specific ratio metrics mentioned earlier. For example, for the disparate impact ratio (DIR) metric:
$$\textit{Parity Deviation}_{DIR}=\left\| 1-\frac{Pr(Y=1\mid V_p=unprivileged)}{Pr(Y=1\mid V_p=privileged)} \right\|$$ 
The ideal deviation lies close to 0.0 (i.e., no deviation from parity), whereas values below 0.2 fall within the acceptable (``fair'') range \cite{bellamy2019ai}. Instead of focusing on individual fairness metrics that assess fairness for specific groups, a fairness meta-metric combines multiple metrics into a single measure to facilitate comparisons between metrics regardless of output range and interpretation.

To ensure that models are both fair and accurate, we employ the AUC-ROC metric and calculate 95\% confidence intervals (CI). Consistent with the benchmark introduction \cite{harutyunyan2019multitask}, our focus on per-instance accuracy leads to calculating overall performance as the micro-average across all predictions, irrespective of the user.

\renewcommand{\arraystretch}{1.1} 
\begin{table}
\caption{\textbf{Datasets used in evaluation}}
\label{tab:datasets}
\resizebox{\columnwidth}{!}{%
\begin{tabular}{lllp{1.8cm}p{2.8cm}p{1.6cm}}
\hline
{\textbf{Data}} & {\textbf{\# Users}} & {\textbf{\# Samples}} & {\textbf{Downstream Task}} & { \textbf{Modalities}}                            & { \textbf{Protected Attributes}} \\
\hline
MIMIC                                & 18.1K & 21.1K                                          & Mortality \newline Prediction                 & Multivariate clinical measurements, e.g, weight, heart rate, blood pressure (M=76)                          & Age, Race, Gender, Language, Insurance           \\
\hline
MESA                                 & 1.8K & 2.2M                                          & Sleep-Wake \newline Classification            & Activity count and white, red, green, blue light measurements (M=5) & Age, Race, Sex                                        \\
\hline
GLOBEM  &    0.7K      &   8.1K    &   Depression Detection   &   Multivariate behavioral signals, e.g., phone use, sleep, location (M=1390)    &   Race, Gender, Disability \\
\hline
\end{tabular}%
}
\vspace{-0.2cm}
\end{table}
\section{Evaluation}\label{sec:evaluation}
We follow the protocol below to assess the applicability and generalizability\footnote{\rev{The term ``generalizability'' refers to the broad applicability of our fairness evaluation framework and findings across multiple real-world datasets and tasks, demonstrating the relative fairness improvements of SSL across diverse data modalities involving human subjects--not the generalization abilities of the models themselves.}} of our framework across datasets.

\noindent\textbf{Overview of Datasets \& Tasks.} Considering the dataset requirements defined in \S\ref{sec:method}, we exclude certain datasets, such as those typically used for fairness research, due to insufficient size for SSL training (e.g., Adult, COMPAS, German Credit), or widely used SSL benchmark datasets, due to modality mismatch (e.g., CelebA, Equity Evaluation Corpus). We also exclude benchmarks used for human-centric tasks such as human-activity recognition due to small sample size, lack of protected attributes, or both (e.g., PAMAP2, MotionSense, UCI-HAR). We select three multimodal, human-centric datasets (Table~\ref{tab:datasets}) spanning the following use cases: in-hospital mortality prediction based on health records and physiological signals, sleep-wake classification based on actigraphy signals, and depression detection based on behavioral signals. \rev{Selected datasets contain different levels of representation bias (Appendix~\ref{ap1}) to evaluate our hypothesis under different scenarios.} For simplicity and readability, we refer to the participating individuals as ``users''.
\begin{enumerate}[label={\arabic*.}, leftmargin=*]
    \item \textbf{MIMIC:} the MIMIC-III Clinical Database \cite{johnson2016mimic} contains more than 31 million clinical events that correspond to 17 clinical variables (e.g., heart rate, oxygen saturation, temperature). 
    Our task involves prediction of in-hospital mortality from observations recorded within 48 hours of an intensive care unit (ICU) admission--a primary outcome of interest in acute care. Following the benchmark preparation workflow by \citet{harutyunyan2019multitask}, we proceed with a total of 18.1K users, forming 21.1K windows, each with 48 timestamps, 76 channels, and no overlap.
    \item \textbf{MESA:} the Multi-Ethnic Study of Atherosclerosis (MESA) \cite{chen2015racial}, contains polysomnography (PSG) and actigraphy data for 1817 out of the initial 2.2K users in the MESA sleep study, based on the benchmark by \citet{palotti2019benchmark}. 
    Our task involves the classification of sleep-wake stages over overnight experiments split into 30-s epochs, forming a total of more than 2.2M windows, each with 101 timestamps, 5 channels, and maximum overlap.
    \item \textbf{GLOBEM:} the multi-year sensing dataset, GLOBEM \cite{xu2022globem_neurips} contains a rich collection of survey and behavioral data, including location, phone usage, physical activity, and sleep, for 497 unique users monitored over four consecutive years for 3-month periods at a time. Our task involves depression detection (self-reported), given a feature matrix including daily feature vectors for the past four weeks. Following the benchmark preparation workflow by \citet{xu2022globem_imwut}, we proceed with a total of more than 8K windows, each with 28 timestamps, and 1390 channels. 
\end{enumerate}

\noindent\textbf{Establishing Protected Attributes.}
Human activities data exhibit variability based on the user's attributes~\cite{spathis2021self}. A starting point for investigating bias is thus to investigate test-time performance for protected attribute groups with different socio-demographic attributes.
Figure~\ref{fig:distributions} (Appendix~\ref{ap1}) shows the (highly imbalanced) distribution of users based on protected attributes. Specifically, the \textit{MIMIC} dataset contains a multitude of protected attributes relevant to the in-hospital mortality task: gender, age, ethnicity, religion, language, and insurance type (a proxy for socioeconomic status). Prior work has revealed disparate treatment in prescribing mechanical ventilation among user groups across ethnicity, gender, and age \cite{meng2022interpretability}, and voiced general fairness concerns for Black and publicly insured users \cite{roosli2022peeking}.
Containing fewer protected attributes, the \textit{MESA} dataset includes age, gender, and ethnicity--highly relevant for sleep classification. Specifically, studies have shown that sleep disorders are more prevalent among older adults, and Black populations and vary with gender and obesity status \cite{chen2015racial}. Finally, \textit{GLOBEM} provides access to gender, race, and disability data upon request.
These attributes are highly relevant for depression prediction, as depression rates are higher in women, people with physical disabilities, and untreated racial minority populations \cite{turner1988physical,bailey2019racial,parker2010gender}.

\noindent\textbf{Training Setup and Hyper-parameter Tuning.}
Following \citet{tang2020exploring}'s recommended architecture for contrastive learning on signals, our model comprises a base encoder featuring three temporal (1D) convolutional layers with kernel sizes of 24, 16, 8, and 32, 64, 96 filters, ReLU activation, a dropout rate of 0.1 (0.4 for GLOBEM), and a concluding global maximum pooling layer. For pre-training, a projection head with three fully-connected layers (256, 128, and 50 units) is utilized, while the fine-tuned evaluation incorporates a classification head with two fully-connected layers (128 and 2 units). Pre-training employs the SGD optimizer with cosine decay of the learning rate over 200 epochs and a batch size of 128. Linear evaluation involves training for 100 epochs with the Adadelta optimizer and a learning rate of 0.03. 

Hyperparameters have been finetuned through grid search across ranges of layer numbers (projection head) [2, 3], batch size [64, 128], epochs with or without early stopping [100, 200], learning rates [0.1, 0.01, 0.03, 0.001] with and without decay [1000, 2000 steps], optimizers [SGD, Adam, Adadelta], and dropout [0.1, 0.3, 0.4, 0.5], and their impact was evaluated on the validation set. \rev{A full grid search would yield over 700 models, but we estimate 100 models conservatively due to untested combinations through pruning.}

\section{Results}\label{sec:results}
To assess the truthfulness of our hypothesis, we explore the impact of fine-tuning in SSL on performance, fairness, and representations.
\begin{figure*}[tbh!]
    \begin{minipage}[]{.35\textwidth}
        \centering
        \includegraphics[width=\textwidth]{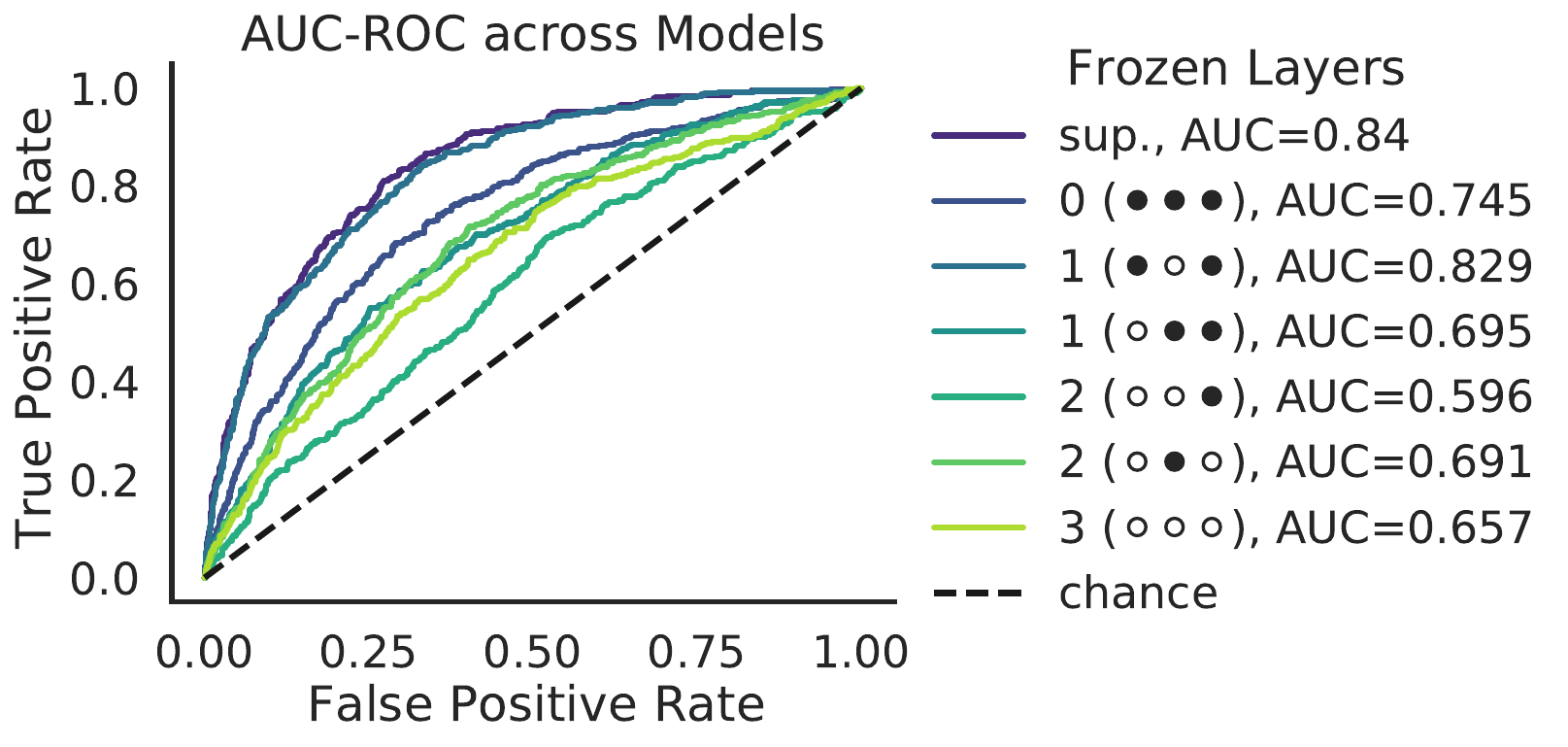}
        \subcaption{MIMIC}\label{fig:auc-mimic}
    \end{minipage}
    \hfill
    \begin{minipage}[]{.32\textwidth}
        \centering
        \includegraphics[width=\textwidth]{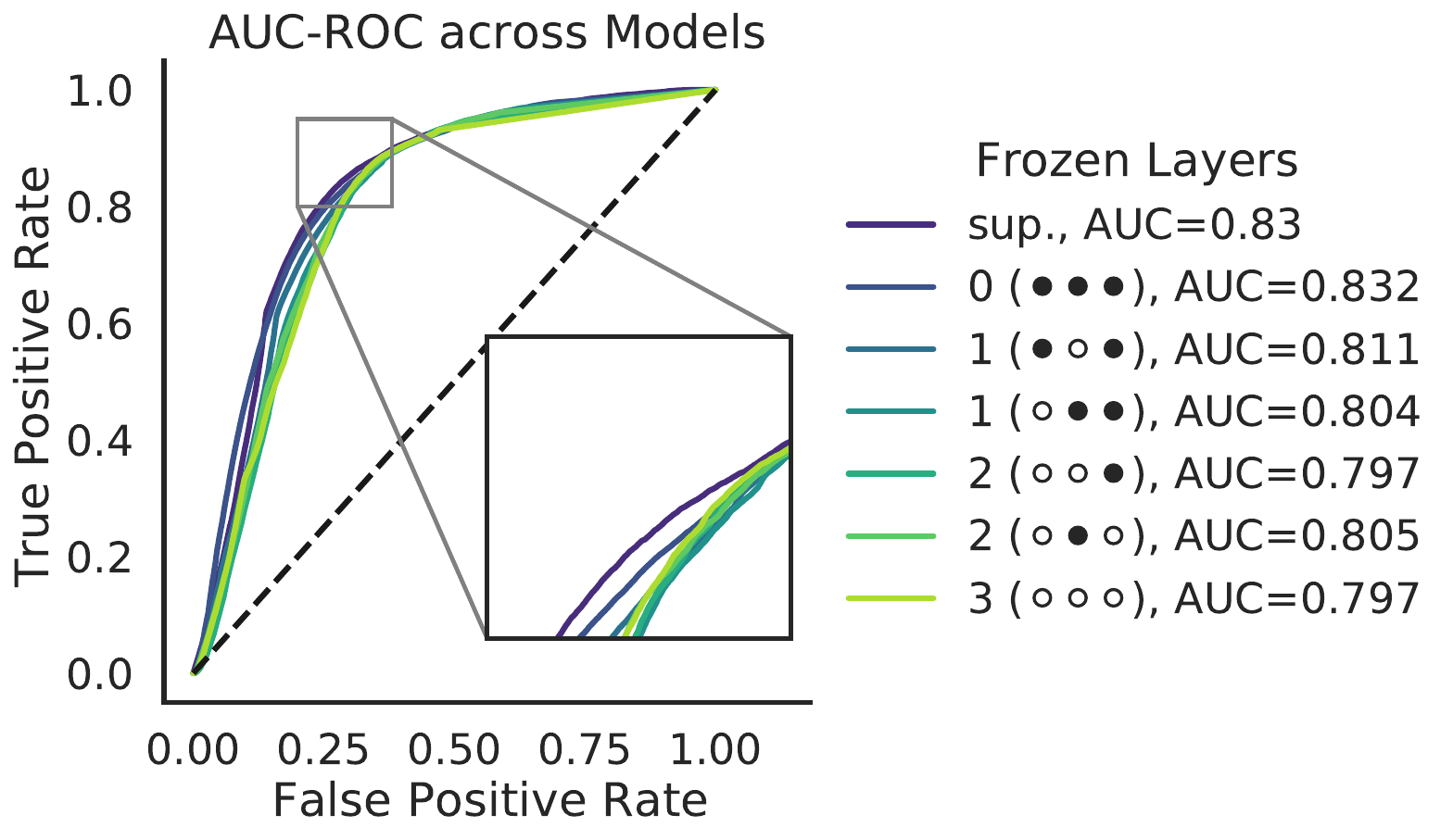}
        \subcaption{MESA}\label{fig:auc-mesa}
    \end{minipage}
    \hfill
    \begin{minipage}[]{.32\textwidth}
        \centering
        \includegraphics[width=\textwidth]{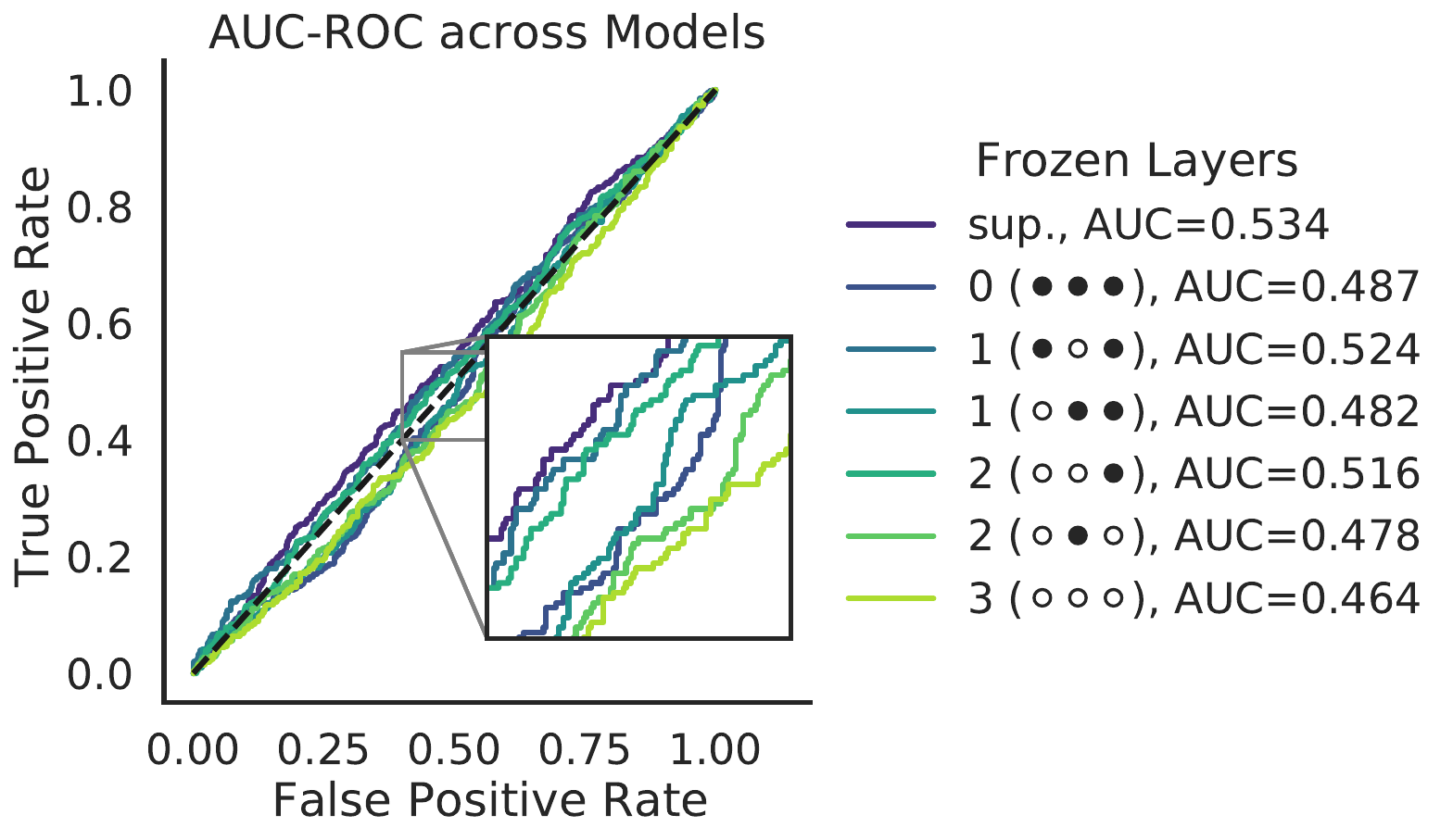}
        \subcaption{GLOBEM}\label{fig:auc-globem}
    \end{minipage}  
    \caption{\textbf{AUC-ROC curves across datasets and fine-tuning strategies.} The supervised models show superior performance, but are closely followed by SSL alternatives, e.g., 1 ($\bullet\circ\bullet$). The level of fine-tuning in SSL greatly affects the observed performance.}
    \label{fig:aucrocscores}
\end{figure*}
\subsection{Impact of Supervision on Performance}\label{sec:results:performance}
\textbf{SSL performs on par with supervised alternatives.} Figure~\ref{fig:aucrocscores} presents the ROC Curves and the AUC-ROC scores for both supervised and SSL models with various levels of fine-tuning across datasets. We notice that for the all datasets, the fully supervised model performs the best in terms of AUC-ROC with a score of 0.84 (CI 0.82-0.86) for MIMIC, 0.83 (CI 0.83-0.83) for MESA, and 0.534 (CI 0.49-0.57) for GLOBEM. In every case, it is closely followed by the SSL model with a single frozen layer (middle) during fine-tuning, i.e., 1 ($\bullet\circ\bullet$) with an AUC-ROC score of 0.829 (CI 0.81-0.85) for MIMIC, 0.811 (CI 0.81-0.81) for MESA, and 0.524 (0.49-0.56) for GLOBEM---a mere 1-2\% loss in overall performance. Such results are in line with prior benchmarking efforts in SSL for visual tasks \cite{newell2020useful} and, closer to our work, human activity recognition \cite{haresamudram2022assessing}.

\renewcommand{\arraystretch}{1.1}
\begin{table*}
\caption{\textbf{Comparison of AUC-ROC between the fully-supervised and the best-performing SSL model, conditioned on protected attributes.} \rev{Numbers in parentheses indicate 95\% Confidence Intervals.} The $\Delta$ columns show differences between a segment and the general population, where yellow indicates disadvantaged and green advantaged segments. The most disadvantaged segment is underlined, and the most advantaged is in bold.} 
\label{tab:aucroctable}
\resizebox{\textwidth}{!}{%
\begin{tabular}{llllllllll}
\hline
 &
   &
   &
  \multicolumn{6}{c}{\textbf{Models}} &
   \\ \hline
\multicolumn{1}{c}{\textbf{Datasets}} &
  \multicolumn{1}{c}{\textbf{Protected Attribute}} &
  \multicolumn{1}{c}{\textbf{Segments}} &
  \multicolumn{2}{l}{\textit{Supervised}} &
  \textit{$\Delta_{segment-general}$} &
  \multicolumn{2}{l}{\textit{SSL (1 $\bullet\circ\bullet$)}} &
  \multicolumn{1}{l|}{\textit{$\Delta_{segment-general}$}} &
  \textit{\textbf{$\Delta_{Supervised-SSL}$}} \\ \hline
 &
  \multicolumn{2}{l}{General Population} &
  \textit{0.839} &
  (0.82-0.86) &
  \textbf{} &
  0.829 &
  (0.81-0.85) &
  \multicolumn{1}{l|}{\textit{}} &
  \textit{} \\ \cline{2-10} 
 &
   &
  $<65$ &
  0.863 &
  (0.83-0.89) &
  \cellcolor[HTML]{F0FAF1}0.024 &
  0.845 &
  (0.8-0.88) &
  \multicolumn{1}{l|}{\cellcolor[HTML]{F5FCF6}0.016} &
  \cellcolor[HTML]{FBFEFC}0.008 \\ \cline{3-10} 
 &
  \multirow{-2}{*}{Age} &
  $\ge65$ &
  0.822 &
  (0.8-0.85) &
  \cellcolor[HTML]{FEFDF0}-0.017 &
  0.82 &
  (0.79-0.85) &
  \multicolumn{1}{l|}{\cellcolor[HTML]{FEFEF6}-0.009} &
  \cellcolor[HTML]{FBFEFC}0.008 \\ \cline{2-10} 
 &
   &
  White &
  0.839 &
  (0.82-0.86) &
  \cellcolor[HTML]{FEFEFD}0 &
  0.831 &
  (0.81-0.86) &
  \multicolumn{1}{l|}{\cellcolor[HTML]{FFFFFF}0.002} &
  \cellcolor[HTML]{FEFEFB}-0.002 \\ \cline{3-10} 
 &
   &
  \underline{Black} &
  \underline{0.762} &
  \underline{(0.65-0.85)} &
  \cellcolor[HTML]{FEF8C2}\underline{-0.077} &
  \underline{0.759} &
  \underline{(0.63-0.85)} &
  \multicolumn{1}{l|}{\cellcolor[HTML]{FEF8C8}\underline{-0.07}} &
  \cellcolor[HTML]{FCFEFC}0.007 \\ \cline{3-10} 
 &
   &
  Asian &
  0.811 &
  (0.68-0.92) &
  \cellcolor[HTML]{FEFCE8}-0.028 &
  0.813 &
  (0.63-0.94) &
  \multicolumn{1}{l|}{\cellcolor[HTML]{FEFDF1}-0.016} &
  \cellcolor[HTML]{F8FDF9}0.012 \\ \cline{3-10} 
 &
  \multirow{-4}{*}{Ethnicity/Race} &
  Hispanic &
  0.955 &
  (0.9-0.99) &
  \cellcolor[HTML]{ADE4B6}0.116 &
  0.937 &
  (0.86-0.98) &
  \multicolumn{1}{l|}{\cellcolor[HTML]{B3E6BB}0.108} &
  \cellcolor[HTML]{FBFEFC}0.008 \\ \cline{2-10} 
 &
   &
  Male &
  0.855 &
  (0.83-0.88) &
  \cellcolor[HTML]{F5FCF6}0.016 &
  0.843 &
  (0.81-0.87) &
  \multicolumn{1}{l|}{\cellcolor[HTML]{F7FDF8}0.014} &
  \cellcolor[HTML]{FFFFFF}0.002 \\ \cline{3-10} 
 &
  \multirow{-2}{*}{Gender} &
  Female &
  0.821 &
  (0.79-0.85) &
  \cellcolor[HTML]{FEFDEF}-0.018 &
  0.812 &
  (0.78-0.84) &
  \multicolumn{1}{l|}{\cellcolor[HTML]{FEFDF0}-0.017} &
  \cellcolor[HTML]{FEFEFE}0.001 \\ \cline{2-10} 
 &
   &
  Medicare &
  0.825 &
  (0.8-0.85) &
  \cellcolor[HTML]{FEFDF2}-0.014 &
  0.819 &
  (0.79-0.84) &
  \multicolumn{1}{l|}{\cellcolor[HTML]{FEFDF5}-0.01} &
  \cellcolor[HTML]{FEFFFE}0.004 \\ \cline{3-10} 
 &
   &
  Private &
  0.868 &
  (0.83-0.9) &
  \cellcolor[HTML]{ECF9EE}0.029 &
  0.856 &
  (0.81-0.9) &
  \multicolumn{1}{l|}{\cellcolor[HTML]{EDF9EF}0.027} &
  \cellcolor[HTML]{FFFFFF}0.002 \\ \cline{3-10} 
 &
   &
  Medicaid &
  0.788 &
  (0.67-0.88) &
  \cellcolor[HTML]{FEFAD6}-0.051 &
  0.786 &
  (0.68-0.87) &
  \multicolumn{1}{l|}{\cellcolor[HTML]{FEFBDC}-0.043} &
  \cellcolor[HTML]{FBFEFC}0.008 \\ \cline{3-10} 
 &
   &
  Government &
  0.885 &
  (0.77-0.99) &
  \cellcolor[HTML]{E0F5E3}0.046 &
  0.895 &
  (0.8-0.98) &
  \multicolumn{1}{l|}{\cellcolor[HTML]{D1F0D6}0.066} &
  \cellcolor[HTML]{FEFDEE}-0.020 \\ \cline{3-10} 
 &
  \multirow{-5}{*}{Insurance} &
  \textbf{Self Pay} &
  \textbf{0.983} &
  \textbf{(0.93-1.0)} &
  \cellcolor[HTML]{99DDA4}\textbf{0.144} &
  \textbf{0.944} &
  \textbf{(0.84-1.0)} &
  \multicolumn{1}{l|}{\cellcolor[HTML]{AEE4B6}\textbf{0.115}} &
  \cellcolor[HTML]{ECF9EE}0.029 \\ \cline{2-10} 
 &
   &
  English &
  0.839 &
  (0.81-0.87) &
  \cellcolor[HTML]{FEFEFD}0 &
  0.831 &
  (0.79-0.86) &
  \multicolumn{1}{l|}{\cellcolor[HTML]{FFFFFF}0.002} &
  \cellcolor[HTML]{FEFEFB}-0.002 \\ \cline{3-10} 
\multirow{-16}{*}{MIMIC} &
  \multirow{-2}{*}{Language} &
  Other &
  0.831 &
  (0.8-0.86) &
  \cellcolor[HTML]{FEFEF7}-0.008 &
  0.82 &
  (0.79-0.84) &
  \multicolumn{1}{l|}{\cellcolor[HTML]{FEFEF6}-0.009} &
  \cellcolor[HTML]{FEFEFC}-0.001 \\ \hline
 &
  \multicolumn{2}{l}{General Population} &
  0.83 &
  (0.83-0.83) &
   &
  0.811 &
  (0.81-0.81) &
  \multicolumn{1}{l|}{} &
   \\ \cline{2-10} 
 &
  Age &
  $\mathbf{<65}$ &
  \textbf{0.856} &
  \textbf{(0.85-0.86)} &
  \cellcolor[HTML]{EDF9EF}\textbf{0.026} &
  \textbf{0.838} &
  \textbf{(0.83-0.84)} &
  \multicolumn{1}{l|}{\cellcolor[HTML]{EDF9EF}\textbf{0.027}} &
  \cellcolor[HTML]{FEFEFD}-0.001 \\ \cline{2-10} 
 &
   &
  $\ge65$ &
  0.813 &
  (0.81-0.81) &
  \cellcolor[HTML]{FEFDF1}-0.017 &
  0.794 &
  (0.79-0.80) &
  \multicolumn{1}{l|}{\cellcolor[HTML]{FEFDF1}-0.017} &
  \cellcolor[HTML]{FEFEFE}0.000 \\ \cline{2-10} 
 &
   &
  White &
  0.838 &
  (0.83-0.84) &
  \cellcolor[HTML]{FAFEFB}0.008 &
  0.82 &
  (0.82-0.82) &
  \multicolumn{1}{l|}{\cellcolor[HTML]{FAFEFA}0.009} &
  \cellcolor[HTML]{FEFEFD}-0.001 \\ \cline{3-10} 
 &
   &
  Black &
  0.833 &
  (0.83-0.84) &
  \cellcolor[HTML]{FEFFFE}0.003 &
  0.808 &
  (0.80-0.81) &
  \multicolumn{1}{l|}{\cellcolor[HTML]{FEFEFB}-0.003} &
  \cellcolor[HTML]{FEFEFE}0.000 \\ \cline{3-10} 
 &
   &
  \underline{Asian} &
  \underline{0.81} &
  \underline{(0.81-0.81)} &
  \cellcolor[HTML]{FEFDEE}\underline{-0.020} &
  \underline{0.8} &
  \underline{(0.8-0.8)} &
  \multicolumn{1}{l|}{\cellcolor[HTML]{FEFDF5}\underline{-0.011}} &
  \cellcolor[HTML]{FAFEFA}0.009 \\ \cline{3-10} 
 &
  \multirow{-4}{*}{Ethnicity/Race} &
  Hispanic &
  0.819 &
  (0.81-0.82) &
  \cellcolor[HTML]{FEFDF5}-0.011 &
  0.801 &
  (0.8-0.8) &
  \multicolumn{1}{l|}{\cellcolor[HTML]{FEFEF6}-0.01} &
  \cellcolor[HTML]{FFFFFF}0.001 \\ \cline{2-10} 
 &
   &
  Male &
  0.831 &
  (0.83-0.83) &
  \cellcolor[HTML]{FFFFFF}0.001 &
  0.813 &
  (0.81-0.82) &
  \multicolumn{1}{l|}{\cellcolor[HTML]{FFFFFF}0.002} &
  \cellcolor[HTML]{FEFEFD}-0.001 \\ \cline{3-10} 
\multirow{-9}{*}{MESA} &
  \multirow{-2}{*}{Gender} &
  Female &
  0.829 &
  (0.82-0.83) &
  \cellcolor[HTML]{FEFEFD}-0.001 &
  0.81 &
  (0.81-0.81) &
  \multicolumn{1}{l|}{\cellcolor[HTML]{FEFEFD}-0.001} &
  \cellcolor[HTML]{FEFEFE}0.000 \\ \hline
 &
  \multicolumn{2}{l}{General Population} &
  0.534 &
  (0.49-0.57) &
   &
  0.524 &
  (0.49-0.56) &
  \multicolumn{1}{l|}{} &
   \\ \cline{2-10} 
 &
   &
  No &
  0.530 &
  (0.5-0.57) &
  \cellcolor[HTML]{FEFDEF}-0.004 &
  0.531 &
  (0.49-0.57) &
  \multicolumn{1}{l|}{\cellcolor[HTML]{FEFEF7}0.007} &
  \cellcolor[HTML]{FEFDF0}-0.003 \\ \cline{3-10} 
 &
  \multirow{-2}{*}{Disability} &
  \underline{Yes} &
  0.613 &
  (0.41-0.8) &
  \cellcolor[HTML]{D0EFD5}.079 &
  \underline{0.323} &
  \underline{(0.17-0.51)} &
  \multicolumn{1}{l|}{\cellcolor[HTML]{FDED61}\underline{-0.201}} &
  \cellcolor[HTML]{FEF39A}-0.122 \\ \cline{2-10} 
 &
   &
  \textbf{White} &
  \textbf{0.651} &
  \textbf{(0.57-0.72)} &
  \cellcolor[HTML]{B2E6BA}\textbf{0.117} &
  0.524 &
  (0.44-0.6) &
  \multicolumn{1}{l|}{\cellcolor[HTML]{FEFDF2}0} &
  \cellcolor[HTML]{B2E6BA}0.117 \\ \cline{3-10} 
 &
   &
  Black &
  N/A &
  N/A &
  N/A &
  N/A &
  N/A &
  \multicolumn{1}{l|}{N/A} &
  N/A \\ \cline{3-10} 
 &
   &
  Asian &
  0.496 &
  (0.45-0.54) &
  \cellcolor[HTML]{FEFAD7}-0.038 &
  0.516 &
  (0.47-0.56) &
  \multicolumn{1}{l|}{\cellcolor[HTML]{FEFDEC}-0.008} &
  \cellcolor[HTML]{F5FCF7}0.030 \\ \cline{3-10} 
 &
   &
  Hispanic/Latinx &
  0.555 &
  (0.41-0.7) &
  \cellcolor[HTML]{FCFEFD}0.021 &
  0.525 &
  (0.39-0.67) &
  \multicolumn{1}{l|}{\cellcolor[HTML]{FEFDF3}0.001} &
  \cellcolor[HTML]{FDFFFD}0.020 \\ \cline{3-10} 
 &
  \multirow{-5}{*}{Ethnicity/Race} &
  Biracial &
  0.449 &
  (0.33-0.56) &
  \cellcolor[HTML]{FEF6B5}-0.085 &
  0.526 &
  (0.43-0.63) &
  \multicolumn{1}{l|}{\cellcolor[HTML]{FEFDF4}0.002} &
  \cellcolor[HTML]{CDEED2}0.083 \\ \cline{2-10} 
 &
   &
  Male &
  0.576 &
  (0.52-0.63) &
  \cellcolor[HTML]{ECF9EE}0.042 &
  0.538 &
  (0.48-0.6) &
  \multicolumn{1}{l|}{\cellcolor[HTML]{FEFEFC}0.014} &
  \cellcolor[HTML]{F7FDF8}0.028 \\ \cline{3-10} 
 &
   &
  Female &
  0.501 &
  (0.45-0.55) &
  \cellcolor[HTML]{FEFBDA}-0.033 &
  0.514 &
  (0.46-0.56) &
  \multicolumn{1}{l|}{\cellcolor[HTML]{FEFCEB}-0.01} &
  \cellcolor[HTML]{FBFEFB}0.023 \\ \cline{3-10} 
 &
   &
  \textbf{Transgender} &
  0.636 &
  (0.33-0.9) &
  \cellcolor[HTML]{BEE9C5}0.102 &
  0.545 &
  (0.2-0.82) &
  \multicolumn{1}{l|}{\cellcolor[HTML]{FCFEFD}0.021} &
  \cellcolor[HTML]{CEEFD3}0.081 \\ \cline{3-10} 
\multirow{-12}{*}{\textit{GLOBEM}} &
  \multirow{-4}{*}{Gender} &
  \underline{\textbf{Other}} &
  \underline{0.25} &
  \underline{(0.0-0.6)} &
  \cellcolor[HTML]{FDE725}\underline{-0.284} &
  \textbf{0.571} &
  \textbf{(0.1-1.0)} &
  \multicolumn{1}{l|}{\cellcolor[HTML]{E8F8EB}\textbf{0.047}} &
  \cellcolor[HTML]{55C667}0.237 \\ \hline
\end{tabular}%
}
\end{table*}
\textbf{Models perform inequitably across protected attributes.} To condition performance on protected attributes, Table~\ref{tab:aucroctable} presents AUC-ROC scores per segment for the supervised and the best-performing self-supervised model. We notice that the models do not perform equitably for all segments. Specifically, for MIMIC, there exists a considerable performance gap experienced by black patients, registering a deviation of nearly $-8\%$ in AUC-ROC, followed by Medicaid-insured patients with deviations exceeding $-5\%$. Conversely, patients with self-insurance show the best performance with deviations up to $+14\%$, trailed by Hispanics with deviations over $+11\%$. 
These findings align with previous studies involving supervised models for MIMIC-III mortality prediction. Notably, Medicaid patients consistently receive inferior predictions despite sharing comparable mortality rates with privately patients. Similarly, black patients consistently underperform compared to white patients, even in the presence of lower mortality rates in the dataset. On the other hand, Hispanic patients exhibit elevated performance attributable to their significantly lower mortality rates compared to other demographic groups \cite{roosli2022peeking} \rev{(for more details on mortality rates see Table~\ref{mortality-rates} in Appendix~\ref{ap1})}. 
Similarly, for GLOBEM, there also exists a significant performance gap experienced by users identifying with gender identities other than the ones included, registering deviations of almost $-30\%$ in AUC-ROC for the supervised model, while users with disabilities register deviations of $-20\%$ for the SSL model. Conversely, White users show the best performance for the supervised model with deviations up to $+12\%$. Lastly, we notice smaller performance discrepancies for MESA, where younger study participants ($<65$ years old) show slightly superior performance ($+3\%$), while Asian participants show slightly declined AUC-ROC scores ($-2\%$). 

\begin{figure}
    \centering
    \includegraphics[width=.5\columnwidth]{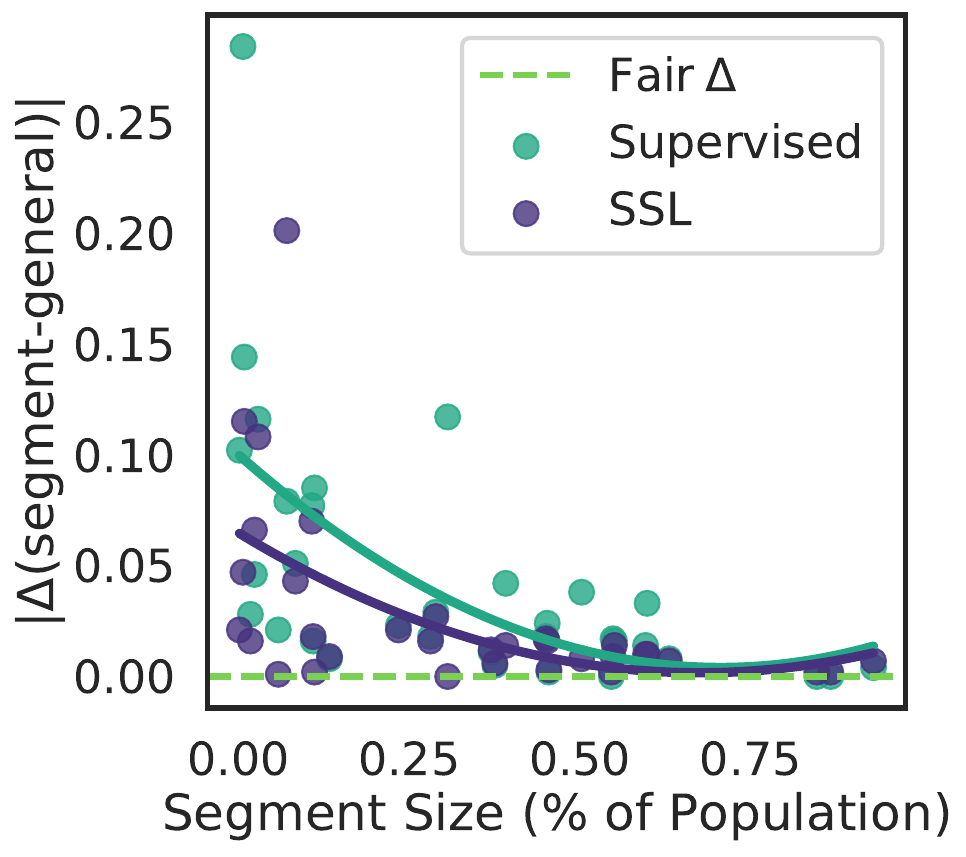}
    \caption{\textbf{The relationship between segment size and performance (AUC-ROC) across datasets. The smaller the segment the larger the performance discrepancies. Fitted lowess curves show that SSL lies closer to the ``fair'' (dashed) line.}} 
    \label{fig:samplesizeVSperformance}
    \vspace{-0.5cm}
\end{figure}
\textbf{SSL is ``fairer'' for smaller segments.} Overall, performance discrepancies are similar between the two models if we focus on the $\Delta_{Supervised-SSL}$ on Table~\ref{tab:aucroctable}, with a slight fairness benefit for the SSL model (indicated by green color). This is more prevalent on the GLOBEM dataset, with an exception of users with disability. However, we should consider the impact of sample size on performance.
Figure~\ref{fig:samplesizeVSperformance} shows the correlation between segment size and performance gap across datasets and protected attributes. The closer the points to the ``fair'' (dashed) line, the smaller the performance gap for this segment with the general population.  For segments $>35\%$ of the population, points are closer to 0.0, whereas smaller segments have much wider gaps. Note that in both cases (i.e., supervised and SSL model), there is a strong negative correlation between segment size and performance gap, namely, the smaller the size, the larger the performance gap. Nevertheless, the lowess curve for the SSL model lies closer to the ``fair'' line, indicating smaller discrepancies between groups.

Nevertheless, performance metrics are not always the best indicator of fairness. Even if a model performs well on average, it might exhibit significant differences in error rates across different groups. For instance, false positives or false negatives may disproportionately affect certain demographic groups, leading to unfair outcomes--a prospect we explore in the following section.

\begin{figure*}
    \begin{minipage}[t]{.35\textwidth}
        \centering
        \includegraphics[width=\textwidth]{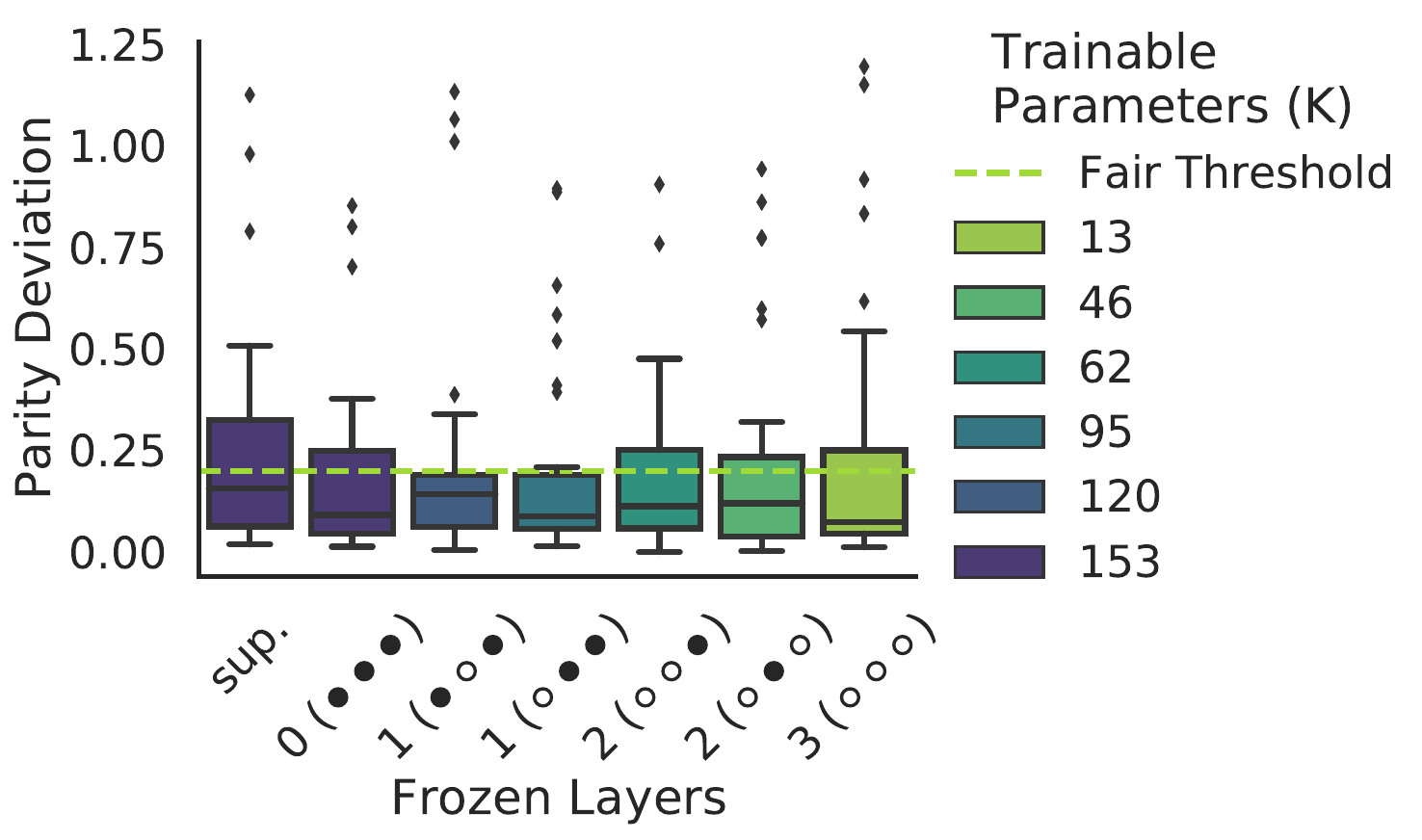}
        \subcaption{MIMIC}\label{fig:fairness-mimic}
    \end{minipage}
    \hfill
    \begin{minipage}[t]{.32\textwidth}
        \centering
        \includegraphics[width=\textwidth]{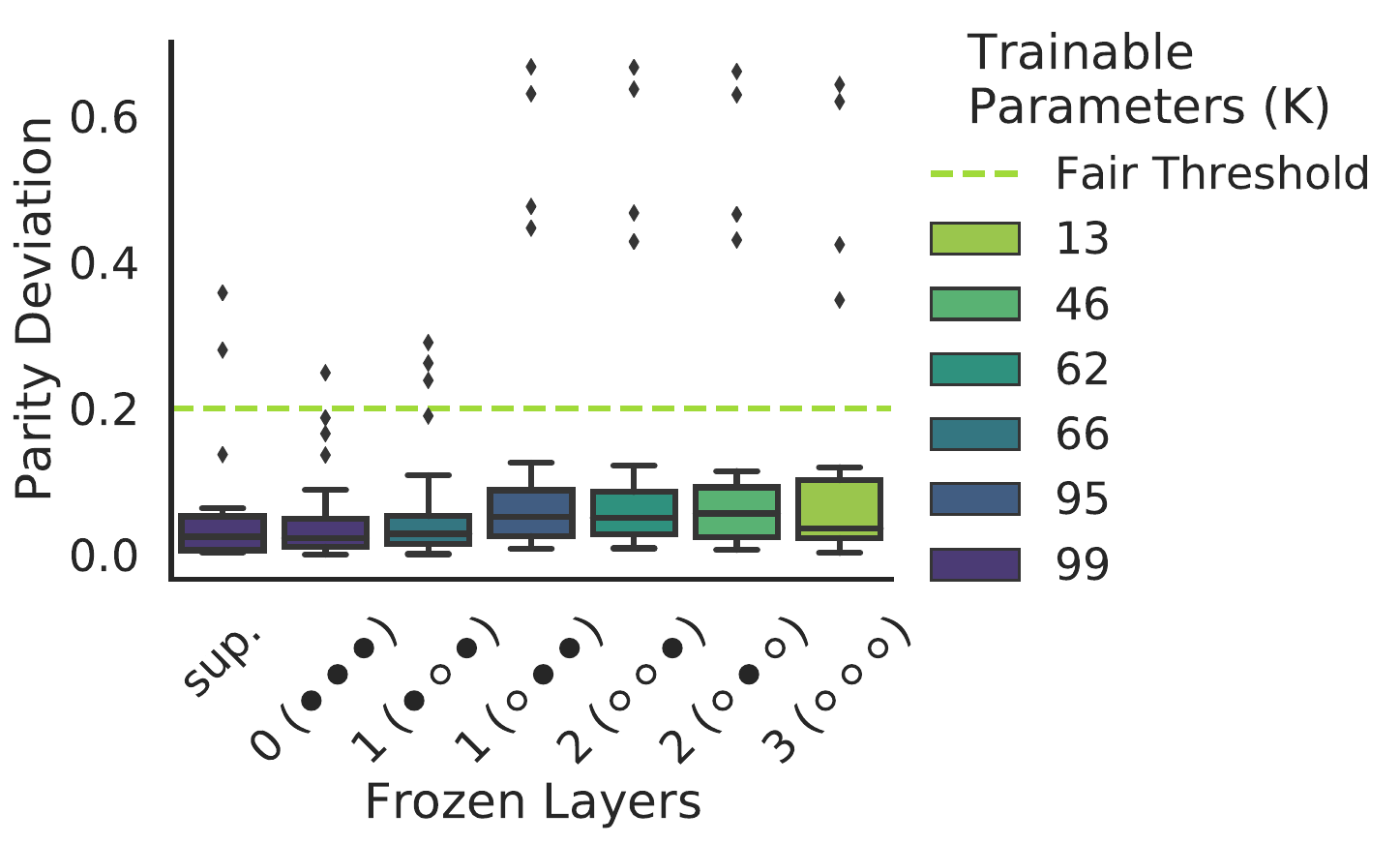}
        \subcaption{MESA}\label{fig:fairness-mesa}
    \end{minipage}
    \hfill
    \begin{minipage}[t]{.32\textwidth}
        \centering
        \includegraphics[width=\textwidth]{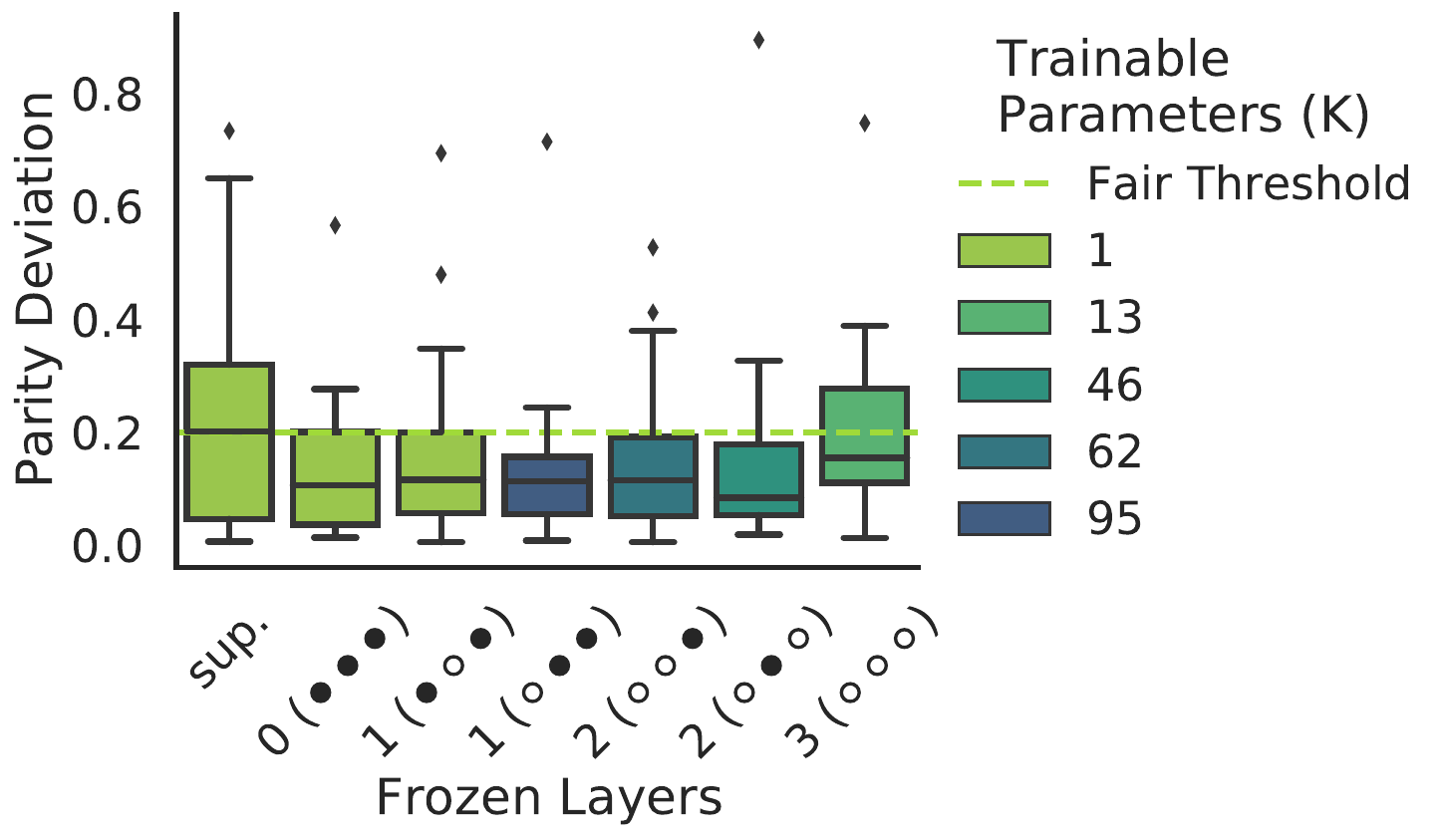}
        \subcaption{GLOBEM}\label{fig:fairness-globem}
    \end{minipage}  
    \caption{\textbf{Relationship between fairness (deviation from parity) and fine-tuning strategies, as a function of model size.} The supervised model has a greater deviation from parity, i.e., increased bias, (dashed line) compared to the best-performing SSL model (i.e., 1 $\bullet\circ\bullet$). The observed ``U-shape'' patterns in MIMIC and GLOBEM datasets suggest an optimal level of fine-tuning.} 
    \label{fig:fairnessscores}
\end{figure*}
\subsection{Impact of Supervision on Fairness}\label{sec:results:fairness}
\textbf{SSL decreases deviation from fairness parity.} Figure~\ref{fig:fairnessscores} shows the deviation of each model's ratio metrics from parity. Deviations greater than 0.2 (dashed line) indicate bias towards a protected attribute, irrespective of privilege. Despite the supervised model having slightly superior performance, it has significantly greater deviation from parity compared to the best-performing SSL model (i.e., 1 $\bullet\circ\bullet$) for the MIMIC and GLOBEM datasets. Specifically, MIMIC's supervised model has on average a 0.24 deviation from parity, while the SSL a 0.21--a 13\% decrease, while GLOBEM's supervised model has a 0.23 deviation from parity, while the SSL a 0.17--a 30\% decrease. 
More importantly, SSL models with a balanced level of unfreezing lie mostly within the acceptable ``fairness'' limits, opposite to the supervised alternative. Note that for the MESA dataset we did not identify any significant differences in parity deviation. This lack of discernible differences could be attributed to the inherent simplicity of the task at hand, i.e., sleep-wake classification, or the more balanced distribution of subjects. \rev{Detailed experimental results on individual fairness metrics comparisons between the SSL model, its linear probing alternative, and the supervised model can be found in Table~\ref{tab:fairness-metrics} on Appendix~\ref{ap:fairness-metrics}}.

\textbf{Middle unfreezing balances performance and fairness.} Additionally, prior work in other domains supports that fine-tuning has an important impact on fairness \cite{ramapuram2021evaluating,rani2023self}. Indeed, our findings illustrate this point for human-centric, multimodal data, too, with statistically significant differences in fairness ratios between SSL models with different levels of fine-tuning (e.g.,  1 $\bullet\circ\bullet$ and 3 $\circ\circ\circ$). This is better illustrated by the observed``U-shape'' patterns in the MIMIC and GLOBEM datasets, suggesting an optimal level of supervision---a sweet spot at middle unfreezing that balances trainable parameters and frozen layers in SSL (Figure~\ref{fig:fairnessscores}).

\begin{figure}
    \centering
    \includegraphics[width=.7\columnwidth]{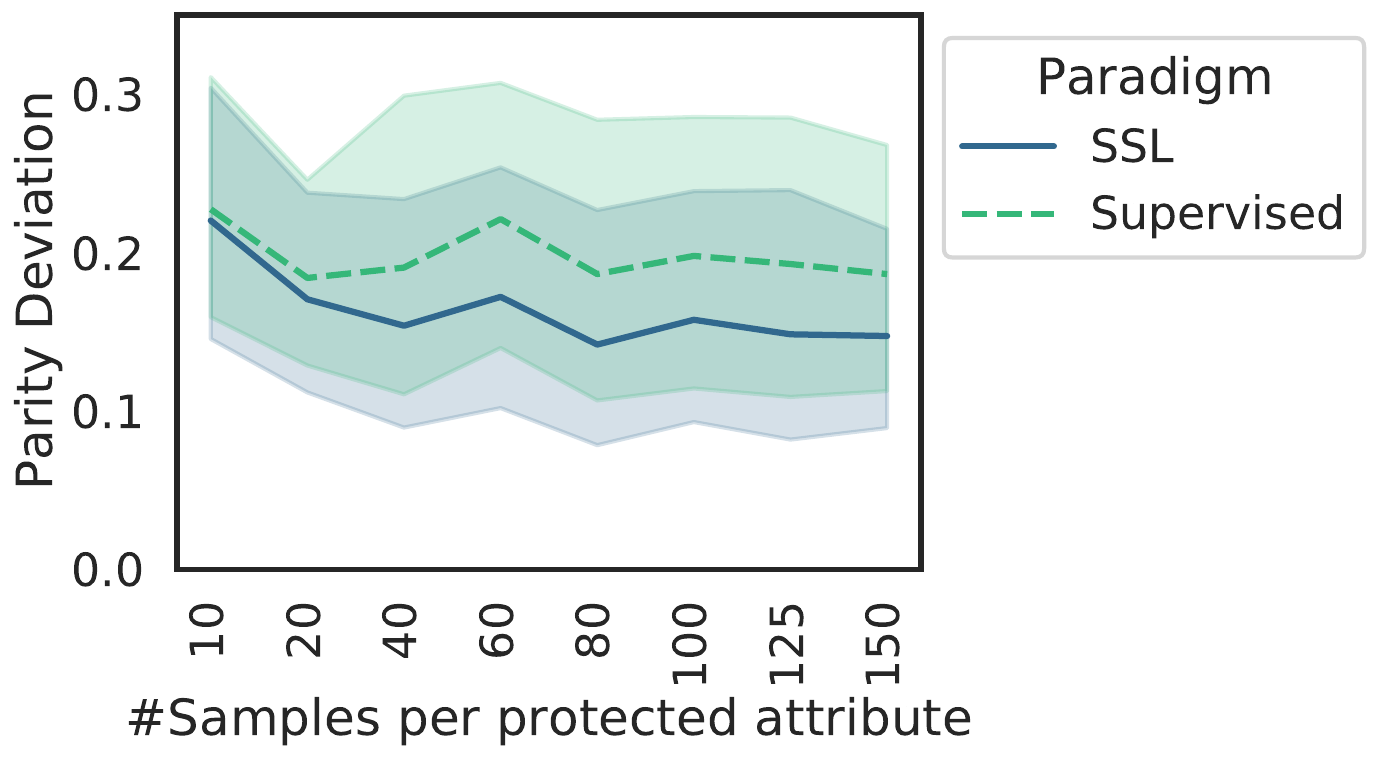}
    \caption{\textbf{Assessing fairness as a function of fine-tuning labelled data in the MIMIC dataset.} The SSL model achieves increased fairness with less training data ($\ge40$ samples per attribute). The shaded error bands represent the range of parity deviation values across fairness metrics and attributes.} 
    \label{fig:dataEfficiency}
    \vspace{-0.5cm}
\end{figure}
\textbf{SSL's fairness gain is more data-efficient.} Following prior work supporting that SSL model can achieve high performance with significantly less training data \cite{tang2021selfhar}, we also assess algorithmic bias (expressed via parity deviation) as a function of limited training data. This evaluation is designed to simulate scenarios where the resources for collecting labeled data are very limited, which might arise in small-scale or academic data collection studies, resulting in limited samples per protected attribute. In this evaluation protocol, a fixed number of labeled samples per ethnicity (i.e., the protected attribute) are extracted from the labeled datasets, and they are the only labeled training data that the models are trained or fine-tuned on. We extract 10-150 samples per ethnicity segment to simulate the different degrees of availability.  Figure~\ref{fig:dataEfficiency} illustrates the parity deviation of models trained on an increasing number of labeled data per protected attribute for the MIMIC dataset, as a case in point. We notice that while the deviation is similar for very limited data ($\le20$), the SSL model shows a quicker fairness gain than the supervised alternative (sample size $\ge40$ per attribute).

\begin{figure}[htb!]
\begin{subfigure}[]{0.27\linewidth}
    \caption*{Random}
    \includegraphics[trim={0 0.4cm 3cm 0},clip,width=\linewidth]{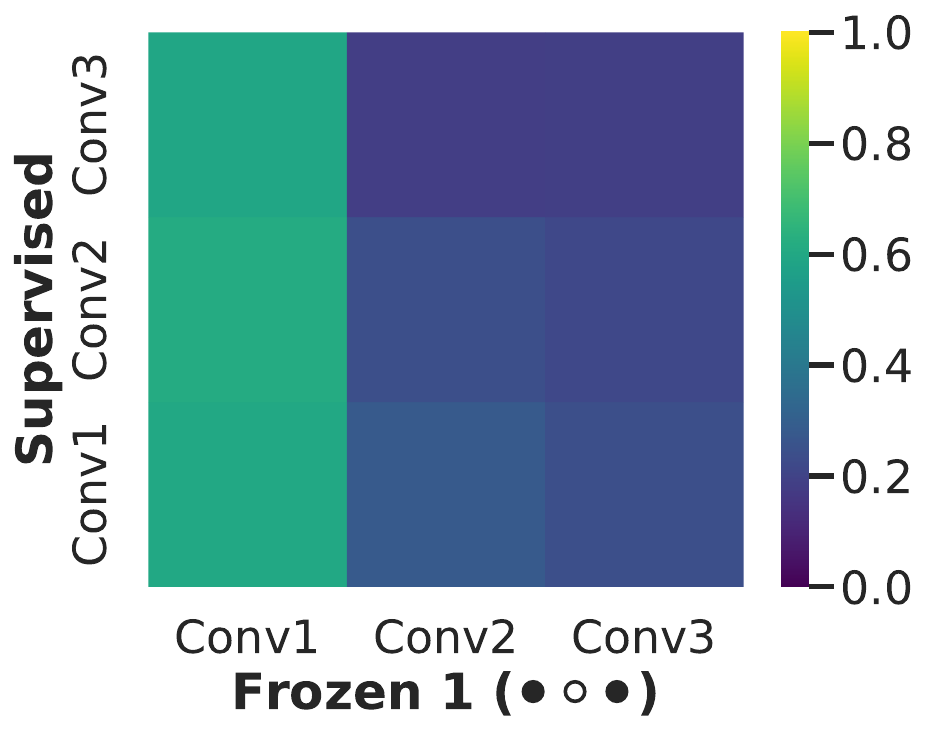}
    \caption{Balanced}
\end{subfigure}
\begin{subfigure}[]{0.21\linewidth}
\caption*{Worst}
    \includegraphics[trim={3cm 0.4cm 3cm 0},clip,width=\linewidth]{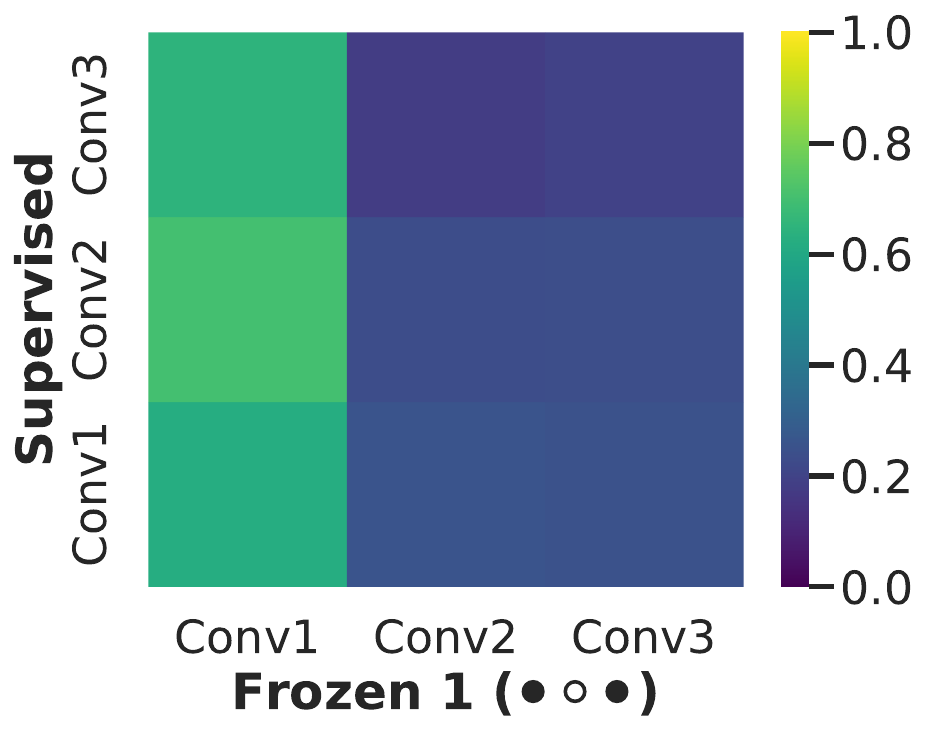}
    \caption{Other}
\end{subfigure} %
\begin{subfigure}[]{0.21\linewidth}
    \caption*{Best}
    \includegraphics[trim={3cm 0.4cm 3cm 0},clip,width=\linewidth]{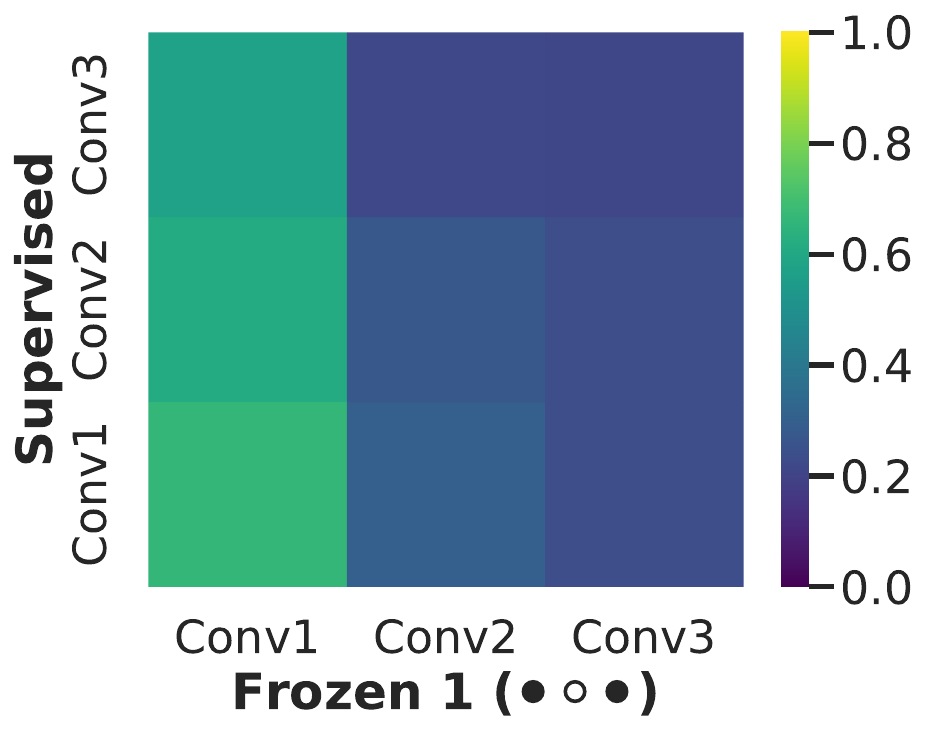}
    \caption{English}
\end{subfigure} %
\begin{subfigure}[]{0.28\linewidth}
    \caption*{Best-Worst}
    \includegraphics[trim={2.5cm 0.4cm 0 0},clip,width=\linewidth]{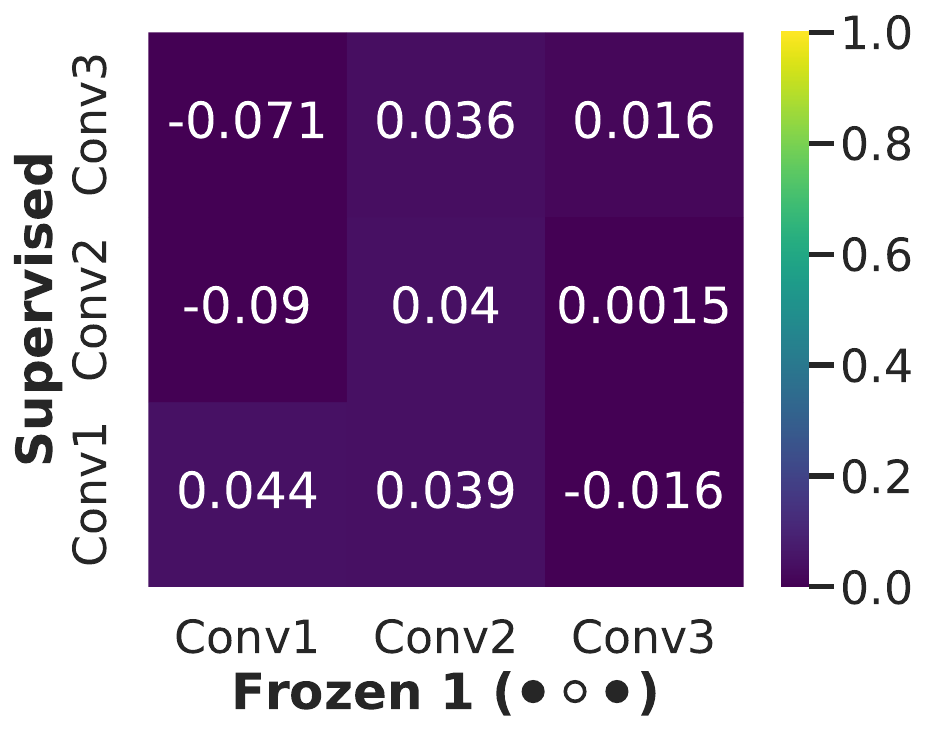}
    \caption{Gap}
\end{subfigure}
\medskip
\begin{subfigure}[]{0.27\linewidth}
    \includegraphics[trim={0 0.4cm 3cm 0},clip,width=\linewidth]{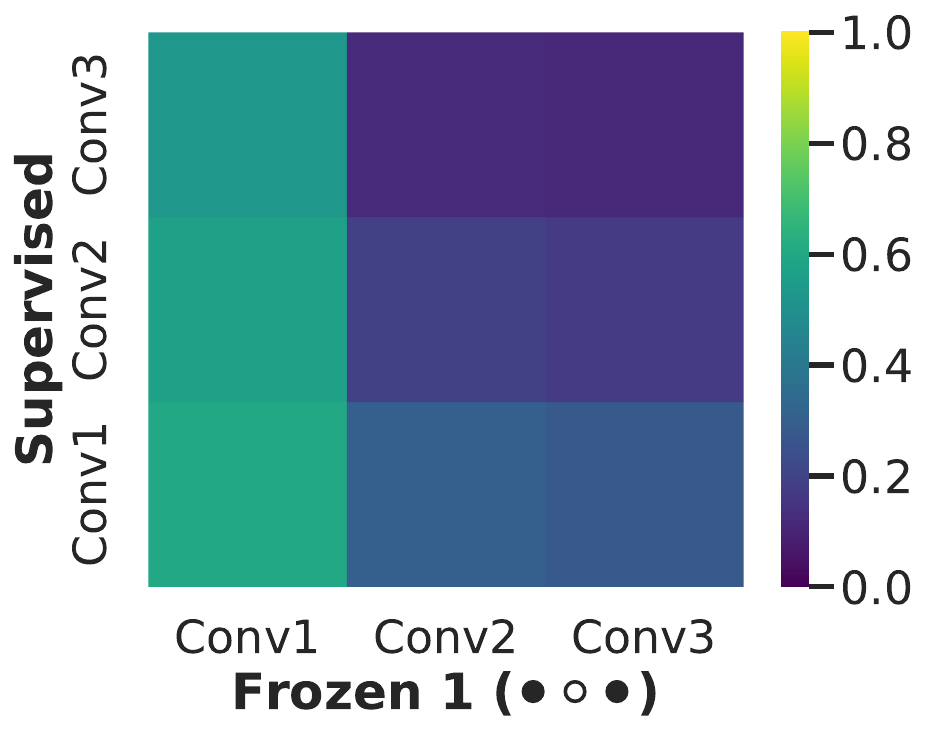}
    \caption{Balanced}
\end{subfigure}
\begin{subfigure}[]{0.21\linewidth}
    \includegraphics[trim={3cm 0.4cm 3cm 0},clip,width=\linewidth]{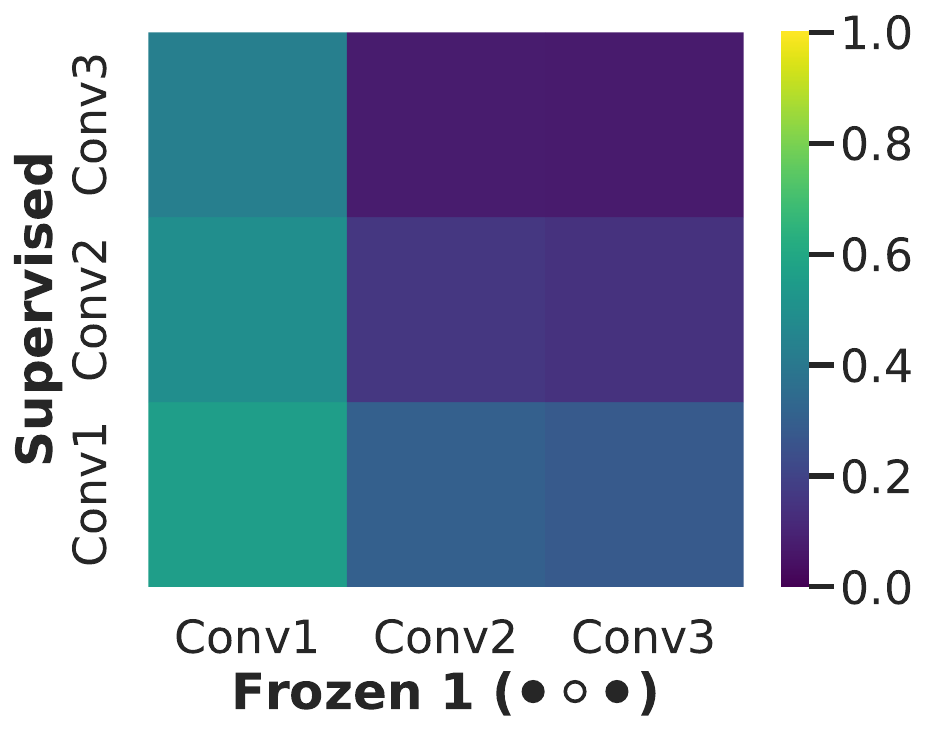}
    \caption{Female}
\end{subfigure} %
\begin{subfigure}[]{0.21\linewidth}
    \includegraphics[trim={3cm 0.4cm 3cm 0},clip,width=\linewidth]{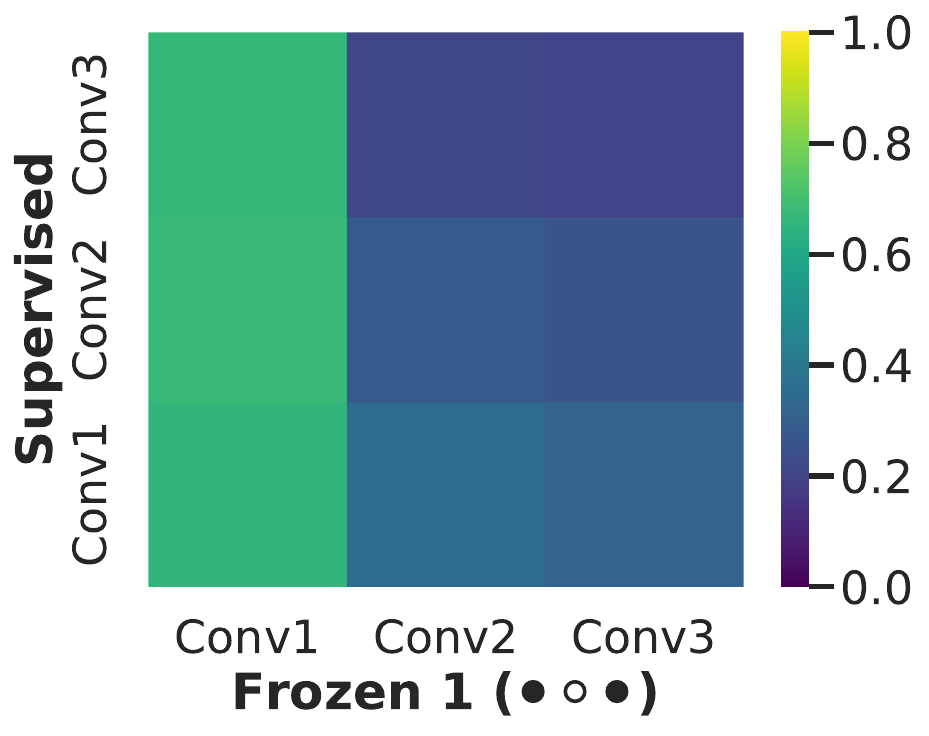}
    \caption{Male}
\end{subfigure} %
\begin{subfigure}[]{0.28\linewidth}
    \includegraphics[trim={2.5cm 0.4cm 0 0},clip,width=\linewidth]{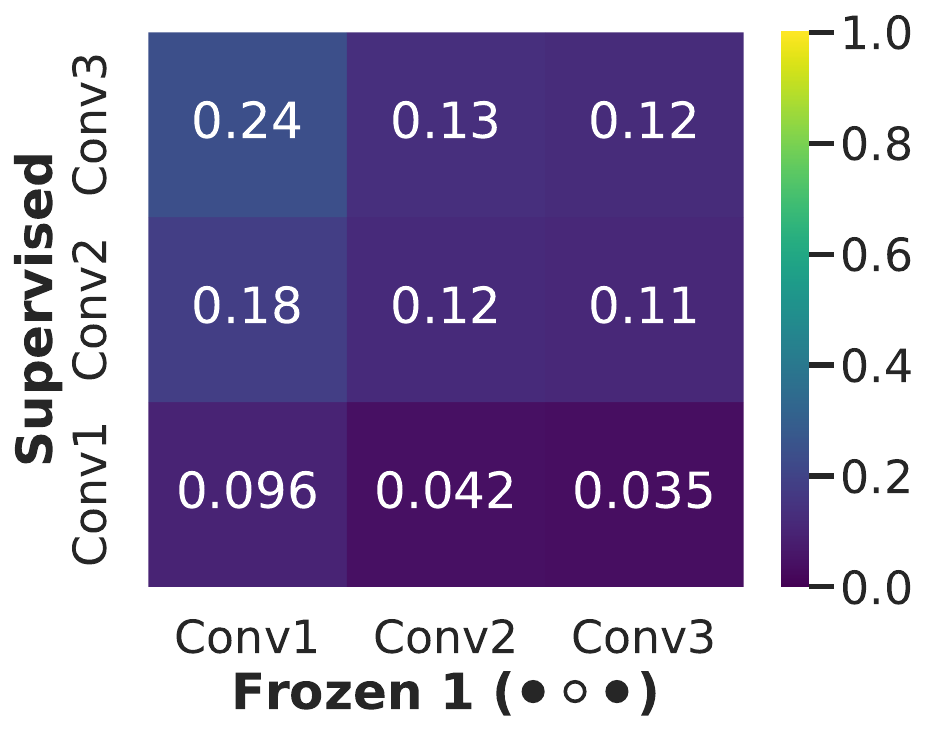}
    \caption{Gap}
\end{subfigure}
\medskip
\begin{subfigure}[]{0.27\linewidth}
    \includegraphics[trim={0 0.4cm 3cm 0},clip,width=\linewidth]{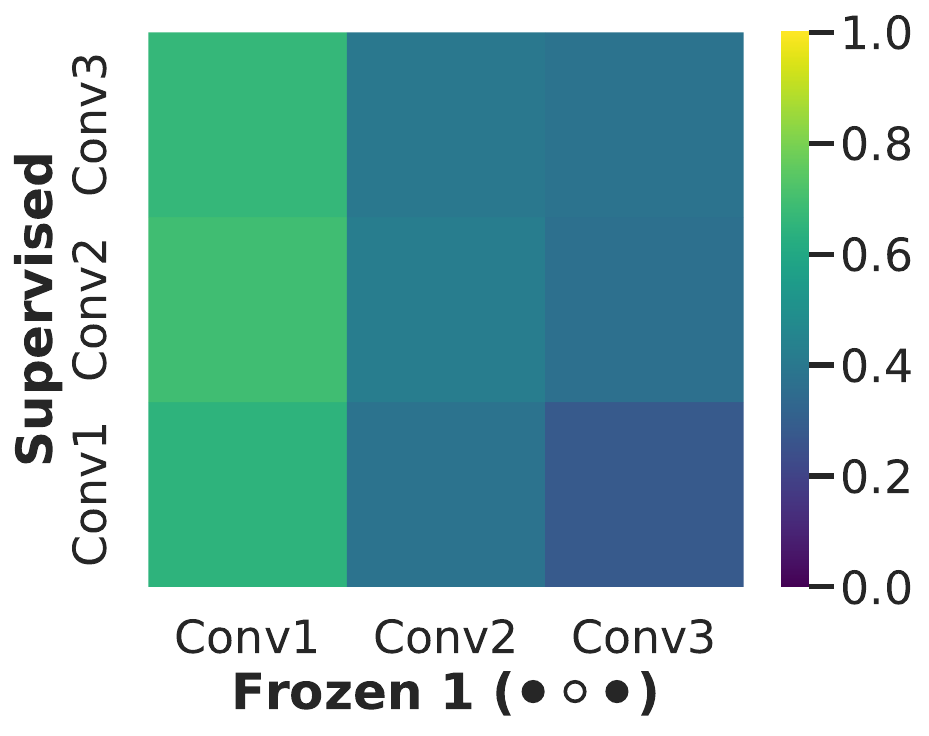}
    \caption{Balanced}
\end{subfigure}
\begin{subfigure}[]{0.21\linewidth}
    \includegraphics[trim={3cm 0.4cm 3cm 0},clip,width=\linewidth]{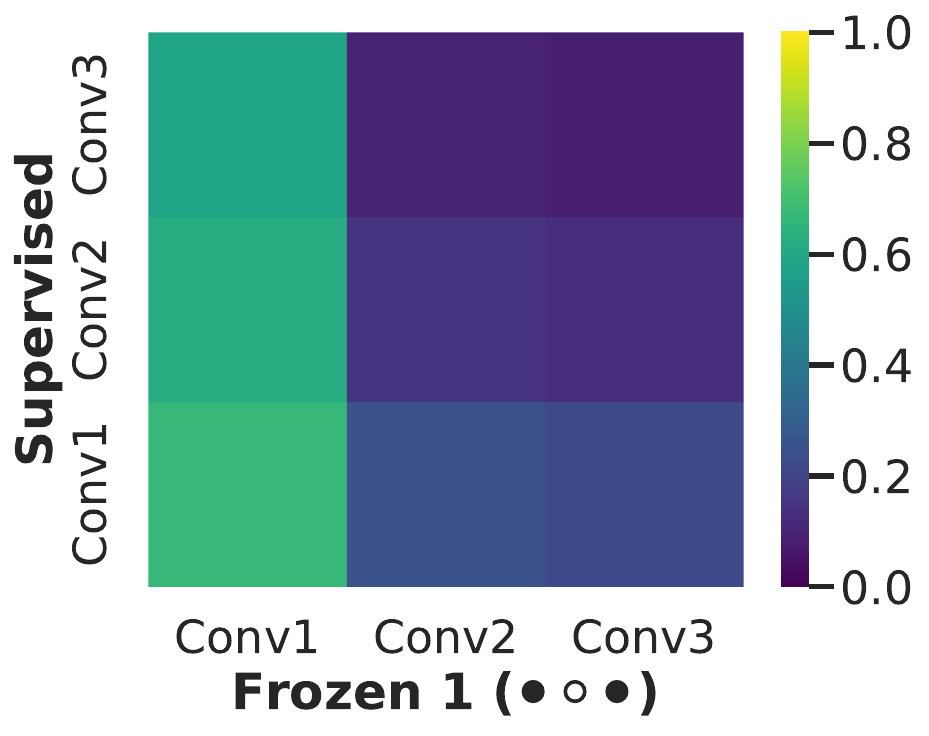}
    \caption{Black}
\end{subfigure} %
\begin{subfigure}[]{0.21\linewidth}
    \includegraphics[trim={3cm 0.4cm 3cm 0},clip,width=\linewidth]{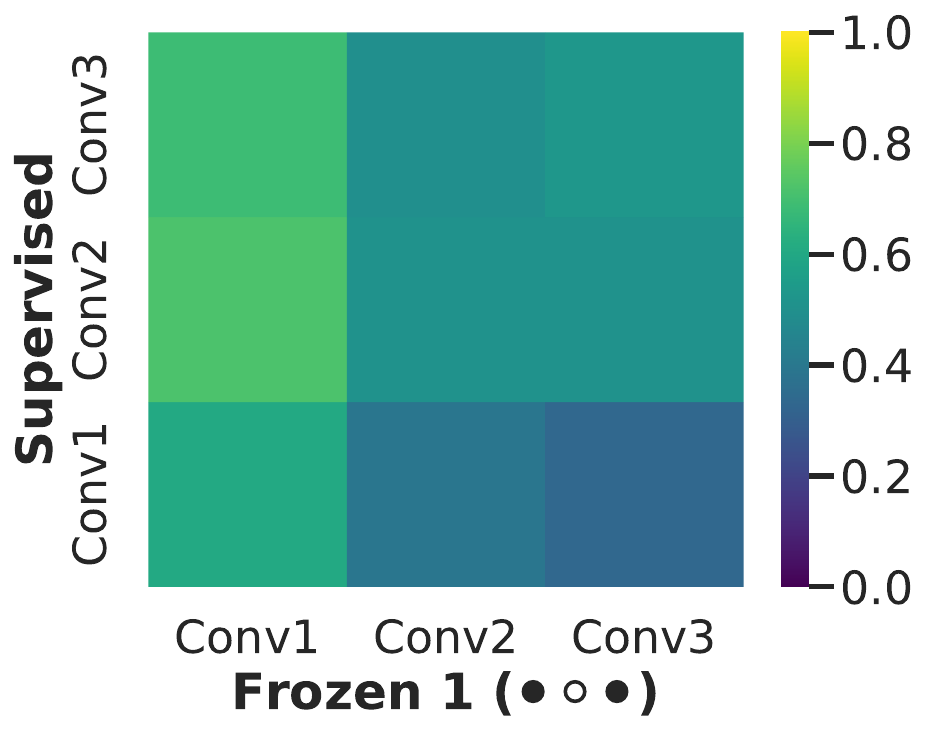}
    \caption{Hispanic}
\end{subfigure} %
\begin{subfigure}[]{0.28\linewidth}
    \includegraphics[trim={2.5cm 0.4cm 0 0},clip,width=\linewidth]{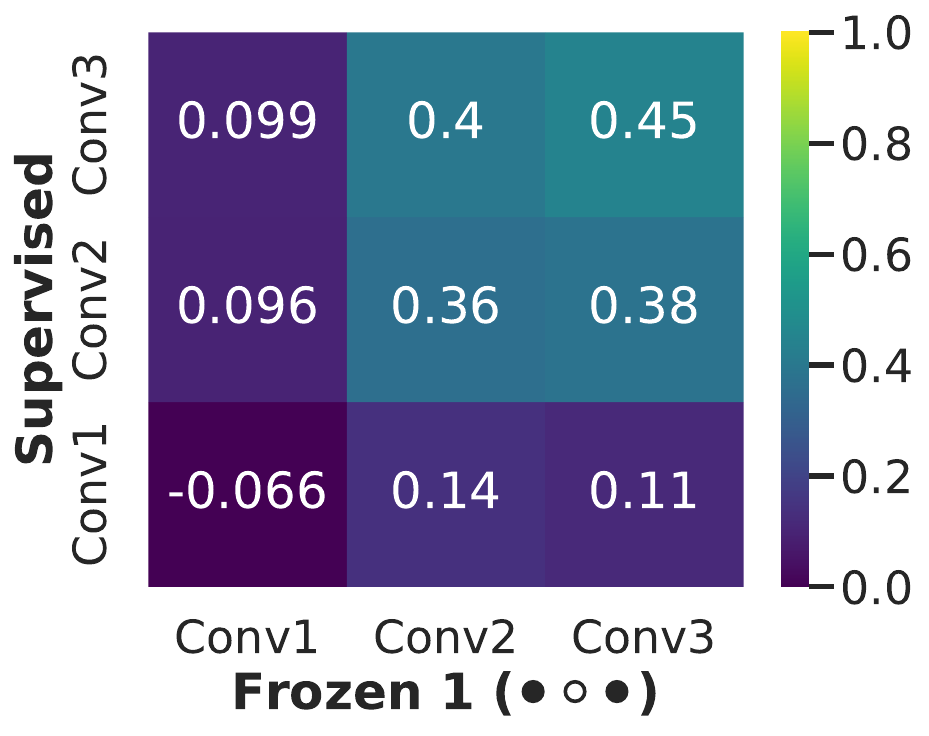}
    \caption{Gap}
\end{subfigure}
\medskip
\begin{subfigure}[]{0.27\linewidth}
    \includegraphics[trim={0 0.4cm 3cm 0},clip,width=\linewidth]{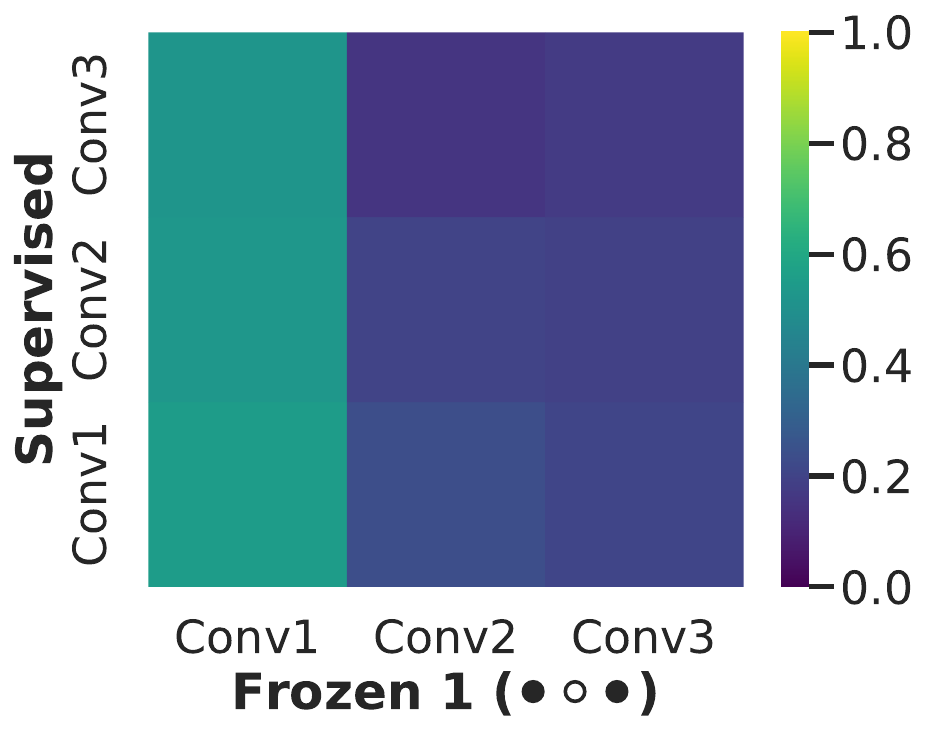}
    \caption{Balanced}
\end{subfigure}
\begin{subfigure}[]{0.21\linewidth}
    \includegraphics[trim={3cm 0.4cm 3cm 0},clip,width=\linewidth]{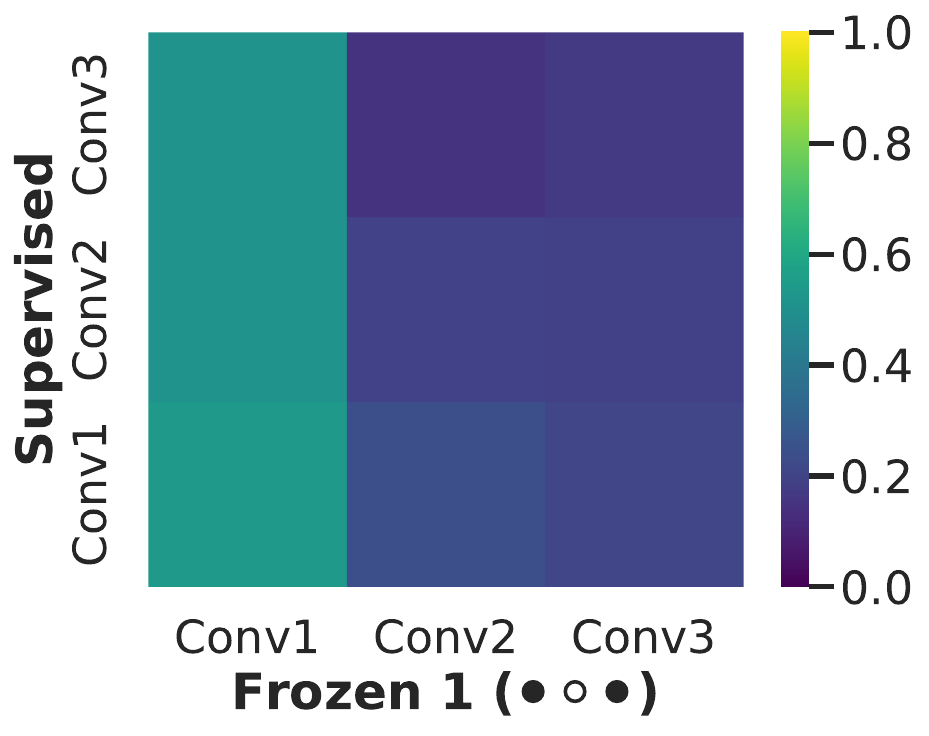}
    \caption{Medicaid}
\end{subfigure} %
\begin{subfigure}[]{0.21\linewidth}
    \includegraphics[trim={3cm 0.4cm 3cm 0},clip,width=\linewidth]{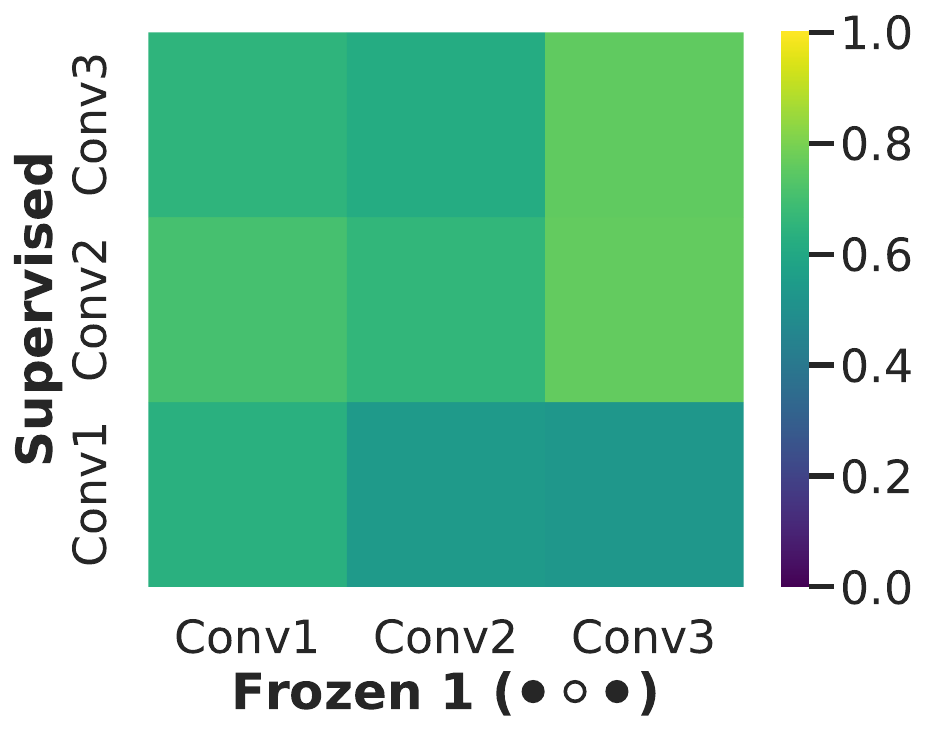}
    \caption{Self Pay}
\end{subfigure} %
\begin{subfigure}[]{0.28\linewidth}
    \includegraphics[trim={2.5cm 0.4cm 0 0},clip,width=\linewidth]{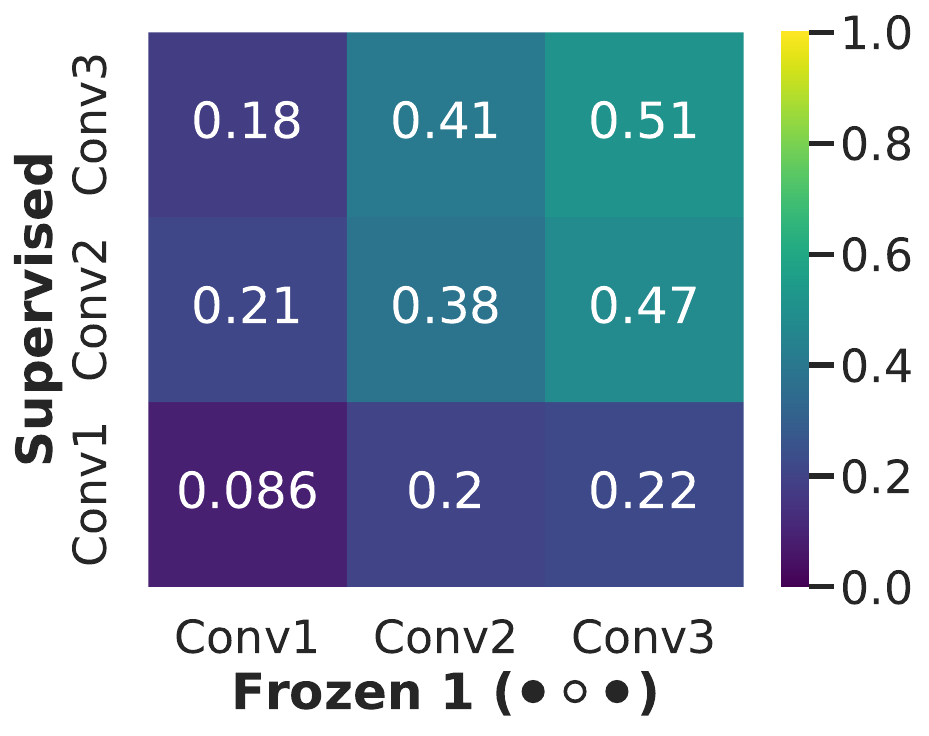}
    \caption{Gap}
\end{subfigure}
\caption{\textbf{Conditioned representation similarity between the supervised and the SSL model through CKA (MIMIC).} Rows correspond to language, gender, race, and insurance. The random subset (first column) is balanced per segment. The similarity is lower for the worst- (second) than the best-performing (third) segment. The higher the performance gap between segments the larger the representation gap (fourth)\label{fig:representationGap}.}
\label{ckaconditionedgender}
\vspace{-0.4cm}
\end{figure}
\subsection{Interplay of Representations and Fairness}\label{sec:results:representation}
\textbf{The larger the performance gap between protected attributes, the greater the fairness deviation.} We compare representation similarity between the supervised and the best-performing SSL model across protected attributes (language, gender, ethnicity, insurance) through CKA. Our findings regarding the impact of supervision on representation learning for time-series data align with prior work on CV. Specifically, \citet{grigg2021self} illustrate how self-supervised and supervised methods learn similar visual representations through dissimilar means and that the learned representations diverge rapidly in the final few layers. Indeed, as illustrated in Figure~\ref{ckaconditionedgender}, the initial layer representations are similar, indicating a shared set of primitives. However, we notice discrepancies at the level of similarity for certain protected attributes. Taking ethnicity as a case in point, we observe a greater similarity in learned representations for Hispanic patients compared to Black patients. This dissimilarity could contribute to the performance gap between these two demographic groups. 
Notably, we find a correlation between the magnitude of the performance gap across patient segments (Table~\ref{tab:minmax}) and the representation similarity gap (Figure~\ref{fig:representationGap}). For instance, the performance gap is minimal for language ($\sim$0\%), slightly larger for gender ($\sim$3\%), and more pronounced for ethnicity and insurance ($\sim$19\%), mirroring the same trend in the similarity gap. For instance, the representation similarity for the best-performing segment is up to $\times13$ greater for the insurance attribute compared to language (median CKA 0.22 to 0.016, respectively).

Interestingly, for the best-performing groups, such as Hispanic or self pay patients, both the SSL and supervised models not only excel in performance but also exhibit strikingly similar representations. This suggests a shared capability in capturing and encoding the underlying data patterns, leading to coherent model outputs. Conversely, when confronted with the worst-performing groups, like Black or Medicaid patients, both models struggle in terms of performance, yet their learned representations diverge notably. Such dissimilarity in their representations implies a focus on different data aspects. The SSL model may be capturing patterns or features not effectively recognized by the supervised model, possibly due to the former's limited reliance on labels. Conversely, the supervised model may emphasize features aligned with labeled data but could face challenges in generalizing due to the inherent complexity of the worst-performing groups (e.g., limited data, class imbalance). A preliminary exploration of the correlation between features learned and protected attributes is given in Appendix~\ref{ap3}. Yet, an in-depth understanding of why the models differ in learned representations for the worst-performing groups is crucial to figuring out the challenges of each learning paradigm in terms of bias.

\section{Discussion \& Conclusions}\label{sec:conclsions}

In conclusion, our investigation into the application of self-supervision in human-centric, multimodal data revealed that SSL models, particularly when fine-tuned with middle unfreezing, can achieve improved fairness compared to supervised models while preserving performance. \rev{Our intuition is that this fine-tuning strategy strikes a balance between retaining knowledge from raw data representations and leveraging information from labeled data.} Interestingly, the SSL's learned representations showcased both similarities and differences with their supervised counterparts, indicating nuanced patterns in capturing and encoding information.

Broadly, the SSL models exhibited smaller deviations from parity across protected attributes, indicating potential effectiveness in mitigating biases associated with downstream labels. \rev{Yet, the focus of this work is on evaluating how design choices in SSL impact fairness, rather than proposing new fairness mitigation algorithms. However, our SSL framework parallels implicit fairness mitigation methods. For instance, the pre-training phase acts akin to pre-processing, removing discriminatory signals by learning from unlabeled data. The subsequent fine-tuning phase operates like an in-processing method, controlling the regularization effect on the model's accuracy. However, it is essential to acknowledge that SSL alone may not eliminate all disparities, especially when trained on poor-quality or biased data, as seen in cases from other domains \cite{omiye2023large}. Future research should explore additional strategies for bias mitigation, and comparative studies with supervised models designed explicitly for bias reduction \cite{liu2021just} are warranted.}

Overall, while SSL presents a positive step towards fairness in real-world, human-centric tasks, it should be considered as part of a broader strategy for addressing bias in ML models, taking into account task-specific nuances and the quality of training data.
The assessment of prediction fairness should consider the data context, and any unfairness arising from insufficient sample sizes or unmeasured predictive variables should be rectified through additional data collection rather than restricting the model.

\begin{acks}
The authors affiliated with the Aristotle University of Thessaloniki acknowledge funding from the European Union’s Horizon 2020 research and innovation programme under the Marie Skłodowska-Curie grant agreement No 813162 \rev{and the Hellenic Artificial Intelligence Society}. The content of this paper reflects only the authors' view and the Agency and the Commission are not responsible for any use that may be made of the information it contains. 
\end{acks}

\bibliographystyle{ACM-Reference-Format}
\bibliography{main}

\appendix
\section{Representation differences across datasets}\label{ap1}
\begin{figure}[hbt!]

\begin{subfigure}{.550\linewidth}
  \includegraphics[width=\linewidth,clip,trim={0 0.1in 2.5in 0}]{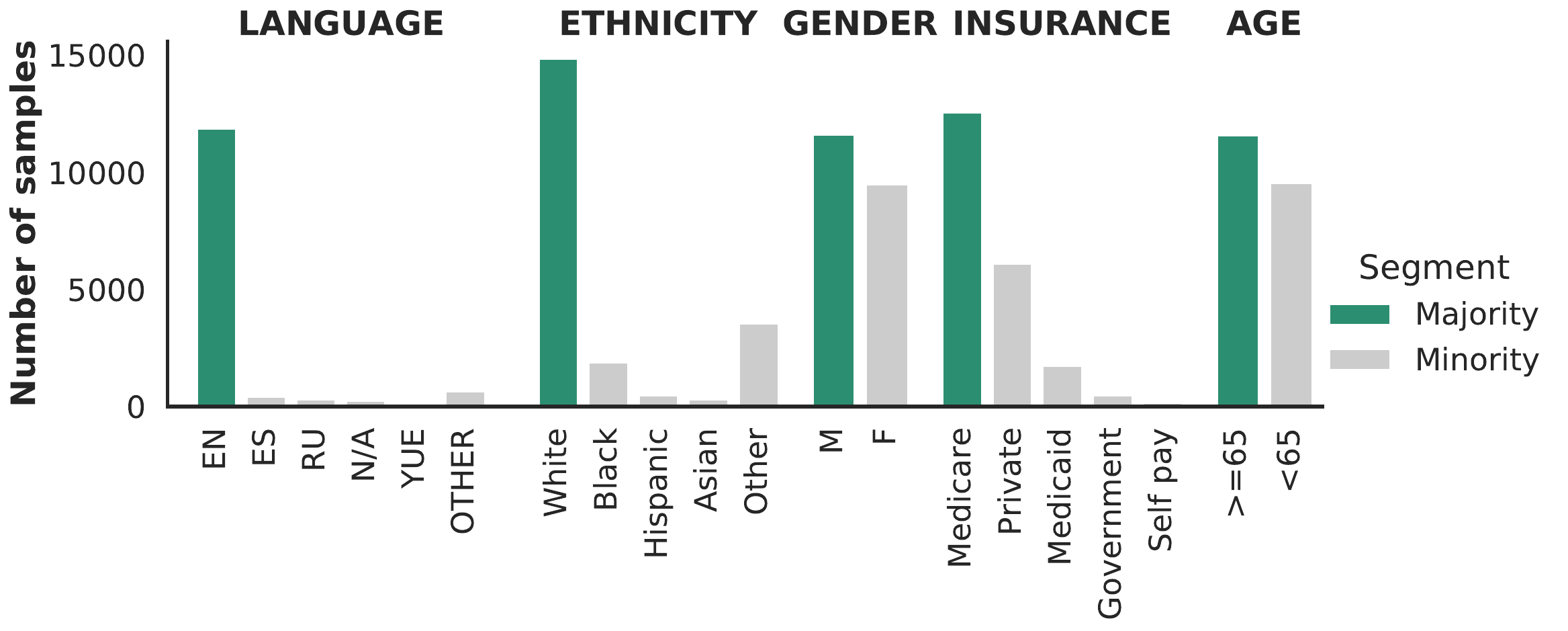}
  \caption{MIMIC Demographics}
  \label{MLEDdet}
\end{subfigure}\hfill %
\begin{subfigure}{.400\linewidth}
  \includegraphics[width=\linewidth]{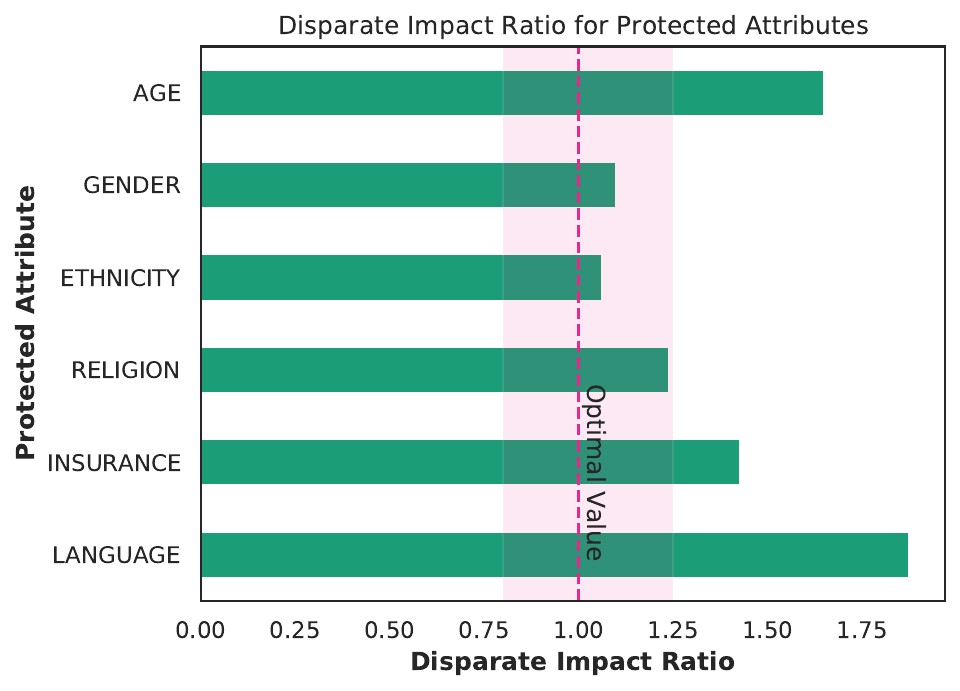}
  \caption{MIMIC Outcome}
  \label{energydetPSK}
\end{subfigure}

\medskip %
\begin{subfigure}{.475\linewidth}
  \includegraphics[width=\linewidth,clip,trim={0 0 2.5in 0}]{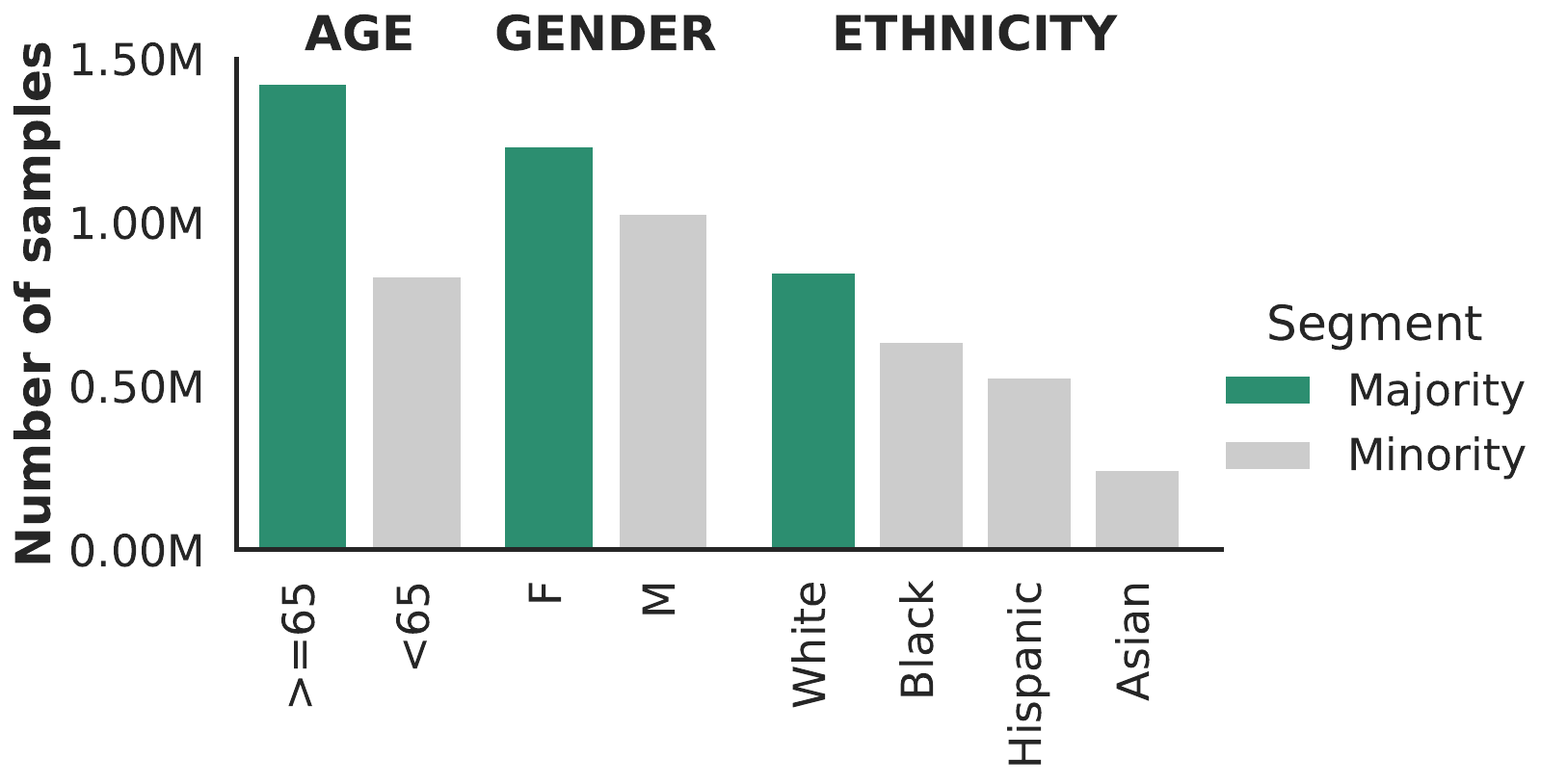}
  \caption{MESA Demographics}
  \label{velcomp}
\end{subfigure}\hfill %
\begin{subfigure}{.475\linewidth}
  \includegraphics[width=\linewidth]{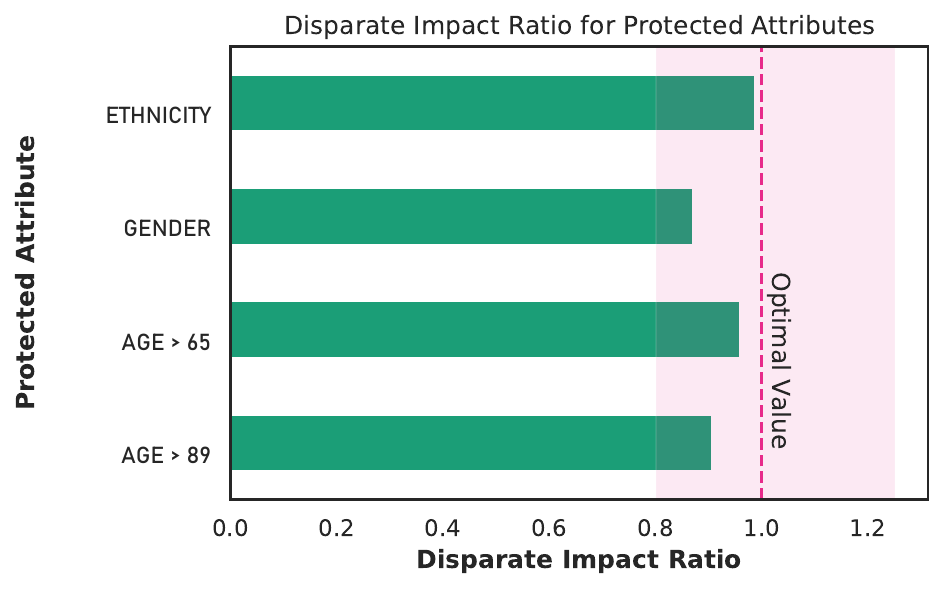}
  \caption{MESA Outcome}
  \label{estcomp}
\end{subfigure}

\medskip %
\begin{subfigure}{.550\linewidth}
  \includegraphics[width=\linewidth]{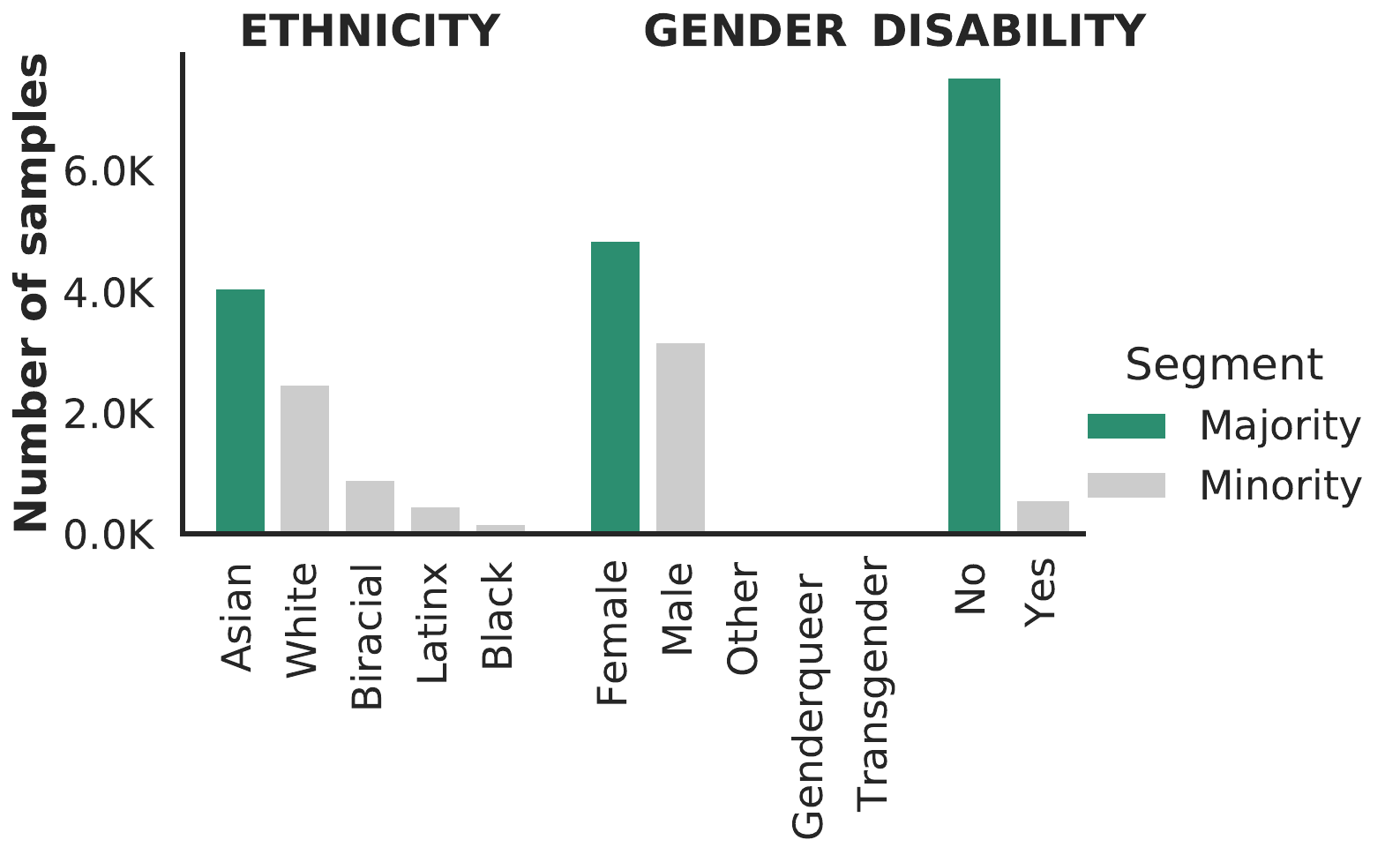}
  \caption{GLOBEM Demographics}
  \label{velcomp}
\end{subfigure}\hfill %
\begin{subfigure}{.400\linewidth}
  \includegraphics[width=\linewidth]{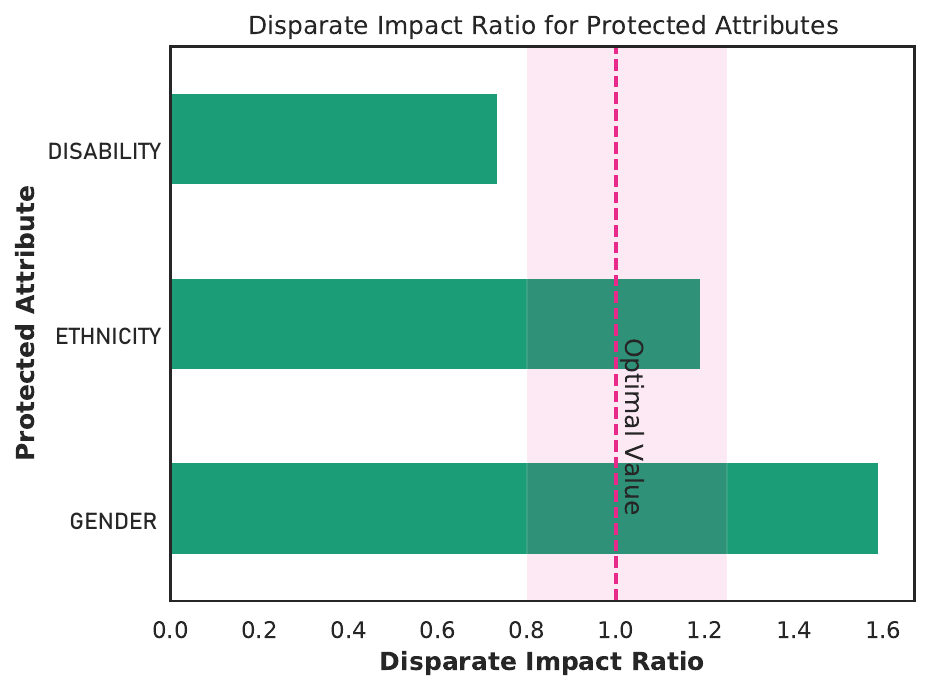}
  \caption{GLOBEM Outcome}
  \label{estcomp}
\end{subfigure}
\caption{\textbf{Distribution of subjects (left) and outcomes (right) based on protected attributes.} MIMIC and GLOBEM datasets are highly imbalanced in terms of demographics \rev{and outcomes -captured via the Disparate Impact Ratio (DIR) metric-} opposite to MESA which shows smaller discrepancies.}
\label{fig:distributions}
\end{figure}

Figure~\ref{fig:distributions} shows the distribution of demographic groups \rev{and outcomes} per dataset. GLOBEM and MIMIC exhibit highly imbalanced distributions, whereas MESA presents a more balanced picture. Notably, in the MIMIC dataset, the majority group constitutes 86.6\% of data points for the language attribute (English speakers), 70.3\% for ethnicity/race (White), 55\% for gender (male), 59.5\% for insurance (Medicare), and 54.8\% for age ($\ge65$). \rev{Simultaneously, demographic groups exhibit distinct mortality rates, as illustrated in Table~\ref{mortality-rates}.} In the GLOBEM dataset, the majority group constitutes 50.2\% for the ethnicity/race attribute (Asian), 59.8\% for gender (Female), and 92.9\% for disability (no disability). Finally, in the MESA dataset, the majority group constitutes 63\% for the age attribute ($\ge65$), 54.5\% for gender (Female), and 37.5\% for race/ethnicity (White). \rev{Regarding outcomes, DIR values outside the shaded region indicate uneven label sampling, which is the case for several attributes in MIMIC and GLOBEM; less so for MESA}. Such representation differences help put our findings into context, as prior work supports that the fairness of predictions should be evaluated in context of the data, and that unfairness can be induced by inadequate samples sizes \cite{chen2018my}.

\begin{table}[]
\caption{\rev{Mortality rates for the MIMIC dataset.}}
\label{mortality-rates}
\begin{tabular}{llll}
\hline
\multicolumn{1}{c}{\textbf{Dataset}} & \multicolumn{1}{c}{\textbf{Protected Attribute}} & \multicolumn{1}{c}{\textbf{Segment}} & \multicolumn{1}{c}{\textbf{Mortality Rate}} \\ \hline
\multirow{15}{*}{MIMIC}              & \multirow{2}{*}{Age}                             & $<65$                                & 9.8\%                                       \\ \cline{3-4} 
                                     &                                                  & $\ge65$                              & 16.0\%                                      \\ \cline{2-4} 
                                     & \multirow{4}{*}{Ethnicity/Race}                  & White                                & 12.9\%                                      \\ \cline{3-4} 
                                     &                                                  & Black                                & 9.2\%                                       \\ \cline{3-4} 
                                     &                                                  & Asian                                & 13.8\%                                      \\ \cline{3-4} 
                                     &                                                  & Hispanic                             & 8.1\%                                       \\ \cline{2-4} 
                                     & \multirow{2}{*}{Gender}                          & Female                               & 13.5\%                                      \\ \cline{3-4} 
                                     &                                                  & Male                                 & 13.0\%                                      \\ \cline{2-4} 
                                     & \multirow{5}{*}{Insurance}                       & Medicare                             & 14.9\%                                      \\ \cline{3-4} 
                                     &                                                  & Private                              & 10.7\%                                      \\ \cline{3-4} 
                                     &                                                  & Medicaid                             & 10.5\%                                      \\ \cline{3-4} 
                                     &                                                  & Government                           & 9.9\%                                       \\ \cline{3-4} 
                                     &                                                  & Self Pay                             & 17.0\%                                      \\ \cline{2-4} 
                                     & \multirow{2}{*}{Language}                        & English                              & 9.9\%                                       \\ \cline{3-4} 
                                     &                                                  & Other                                & 17.5\%                                      \\ \hline
\end{tabular}
\vspace{-0.5cm}
\end{table}

\section{Intra- and Inter-group distances in intermediate representations}\label{ap3}
\begin{figure}[tbh!]
    \centering
    \includegraphics[width=\columnwidth]{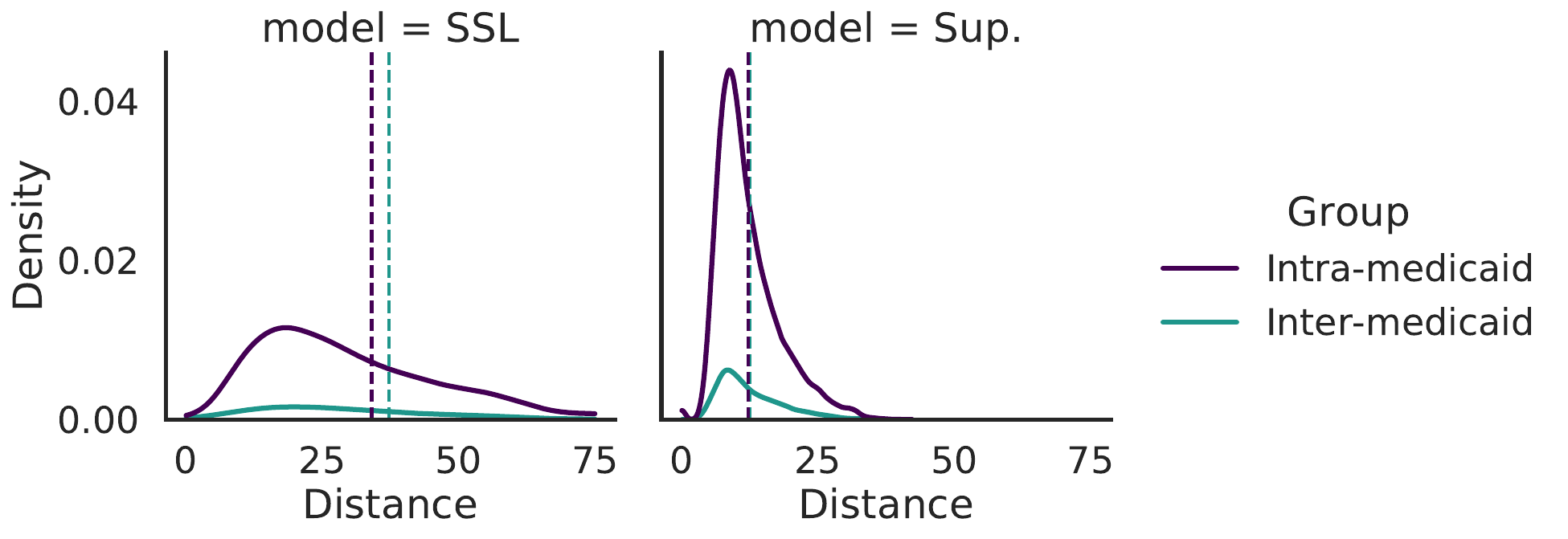}
    \caption{\textbf{Distribution of intra-group and inter-group distances in intermediate representations between the best-performing and the worst-performing segment.} Dashed lines represent the group mean. In the SSL model, intra-distances within the worst-performing segment are smaller compared to inter-distances with the best-performing segments.}
    \label{fig:pairwise}
\end{figure}

 To investigate the correlation between features learned by the SSL model and protected attributes, we first determine the medoids, representing the most representative patients, for each demographic segment, and then, we compute the average distances between these medoids. Within the SSL segments, we observe a significant increase in separability, with distances being 70\% larger on average ($L1\mhyphen norm_{sup}=4.32$, $L1\mhyphen norm_{ssl}=7.34$). This implies that SSL's decision-making process is, in part, influenced by representations specific to protected attributes. This tendency is further illustrated in Figure~\ref{fig:pairwise}, using the insurance attribute as a case in point. Specifically, in the SSL model, intra-distances within the worst-performing segment (Medicaid patients) are smaller than inter-distances between the worst-performing and the best-performing (self pay patients) segments. Such a distinctive pattern is notably absent in the supervised model, emphasizing the potential role of protected-attribute-specific representations in SSL's learning process. For comparison, Table~\ref{tab:minmax} illustrates the performance discrepancy between the worst- and best-performing group (in AUC-ROC), e.g., Medicaid \textit{vs.} self-pay for the insurance attribute.
\vspace{-0.4cm}

\begin{table}[htb!]
\caption{\rev{The disparity between the best and worst-performing groups for the MIMIC dataset is smaller for the SSL model.\label{tab:minmax}}}
\begin{tabular}{llll}
\hline
\multicolumn{1}{c}{\multirow{2}{*}{\textbf{Dataset}}} & \multicolumn{1}{c}{\multirow{2}{*}{\textbf{Protected Attribute}}} & \multicolumn{2}{c}{\textbf{Model}} \\ \cline{3-4} 
\multicolumn{1}{c}{}                                  & \multicolumn{1}{c}{}                                              & \textit{Sup}     & \textit{SSL}    \\ \hline
\multirow{5}{*}{MIMIC}                                & Age                                                               & 0.04             & \textbf{0.03}   \\ \cline{2-4} 
                                                      & Race                                                              & 0.19             & \textbf{0.18}   \\ \cline{2-4} 
                                                      & Gender                                                            & 0.03             & \textbf{0.03}   \\ \cline{2-4} 
                                                      & Insurance                                                         & 0.20             & \textbf{0.16}   \\ \cline{2-4} 
                                                      & Language                                                          & 0.01             & 0.01            \\ \hline
\end{tabular}
\end{table}
\section{Fairness Metrics}\label{ap:fairness-metrics}
\rev{Apart from the AUC-ROC performance metric, we utilize six popular fairness metrics for our evaluation:}
\begin{itemize}
    \item \rev{Disparate Impact Ratio (DIR): Ratio of selection rates.}
    $$\frac{Pr(\hat{Y} = 1 | D = \text{unprivileged})}
  {Pr(\hat{Y} = 1 | D = \text{privileged})}$$
  \item \rev{False Discovery Rate Ratio (FDR): Ratio of the proportion of false positives (incorrectly predicted positive cases) to the number of total positive results between different demographic groups, providing a measure to evaluate disparities in model errors across those groups.}
  $$\frac{FDR_{D = \text{unprivileged}}}{FDR_{D = \text{privileged}}} \text{ where } FDR = FP/(TP + FP)$$
  \item \rev{False Negative Rate Ratio (FNR): Ratio of the proportion of actual positive cases incorrectly predicted as negative between different demographic groups, serving as a measure to assess disparities in model performance across those groups.}
  $$\frac{FNR_{D = \text{unprivileged}}}{FNR_{D = \text{privileged}}} \text{ where } FNR = FN/P$$
  \item \rev{False Omission Rate Ratio (FOR): Ratio of the proportion of false negatives to the number of total negative results between different demographic groups, offering a metric to assess disparities in model omissions across those groups.}
  $$\frac{FOR_{D = \text{unprivileged}}}{FOR_{D = \text{privileged}}} \text{ where } FOR = FN/(TN + FN)$$
  \item \rev{False Positive Rate Ratio (FPR): Ratio of the proportion of false positives (incorrectly predicted positive cases) between different demographic groups, serving as a metric to evaluate disparities in model errors across those groups.}
  $$\frac{FPR_{D = \text{unprivileged}}}{FPR_{D = \text{privileged}}} \text{ where } FPR = FP/N$$
\end{itemize}

\rev{We utilize the parity deviation meta-metric (Section~\ref{sec:method}), as a means to facilitate the comparison between multiple ratio-based fairness metrics. The usage of multiple fairness metrics, (e.g., false positive rate ratio, false negative rate ratio, etc.) in the healthcare and well-being setting is not uncommon \cite{pias2023undersampling,sogancioglu2022effects}. On the contrary, the usage of multiple metrics is recommended, to capture diverse fairness perspectives in human-centric applications \cite{yfantidou2023beyond}.}
\begin{table}[]
\caption{\rev{Fairness metrics by dataset and protected attribute. Values outside the accepted range (>=0.2 parity deviation) are colored in purple. For those cases, the SSL model performs better or same to the supervised (Sup.) or linear probing (LP) model for all protected attributes in MIMIC and GLOBEM.}}
\label{tab:fairness-metrics}
\resizebox{\columnwidth}{!}{%
\begin{tabular}{llllllll}
\hline
\multicolumn{1}{c}{\textbf{Dataset}} &
  \multicolumn{1}{c}{\textbf{Protected Attribute}} &
  \multicolumn{1}{c}{\textbf{Model}} &
  \multicolumn{1}{c}{\textbf{DIR}} &
  \multicolumn{1}{c}{\textbf{FDR}} &
  \multicolumn{1}{c}{\textbf{FNR}} &
  \multicolumn{1}{c}{\textbf{FOR}} &
  \multicolumn{1}{c}{\textbf{FPR}} \\ \hline
 &
   &
  SSL &
  1.19391 &
  0.897846 &
  1.187938 &
  \cellcolor[HTML]{CBCEFB}2.132436 &
  1.142868 \\ \cline{3-8} 
 &
   &
  Sup. &
  \cellcolor[HTML]{CBCEFB}1.34903 &
  0.917276 &
  1.093168 &
  \cellcolor[HTML]{CBCEFB}1.979175 &
  \cellcolor[HTML]{CBCEFB}1.319301 \\ \cline{3-8} 
 &
  \multirow{-3}{*}{Age} &
  LP &
  1.047373 &
  0.918621 &
  1.083654 &
  \cellcolor[HTML]{CBCEFB}1.832518 &
  1.025794 \\ \cline{2-8} 
 &
   &
  SSL &
  1.078457 &
  0.977301 &
  0.847877 &
  0.93395 &
  1.063312 \\ \cline{3-8} 
 &
   &
  Sup. &
  1.029786 &
  0.940589 &
  0.877699 &
  0.945341 &
  0.977184 \\ \cline{3-8} 
 &
  \multirow{-3}{*}{Ethnicity} &
  LP &
  0.984655 &
  0.964173 &
  0.982594 &
  1.047049 &
  0.957786 \\ \cline{2-8} 
 &
   &
  SSL &
  1.156379 &
  1.005544 &
  0.959478 &
  1.140239 &
  1.180523 \\ \cline{3-8} 
 &
   &
  Sup. &
  \cellcolor[HTML]{CBCEFB}1.26365 &
  1.04365 &
  0.934915 &
  1.116688 &
  \cellcolor[HTML]{CBCEFB}1.338922 \\ \cline{3-8} 
 &
  \multirow{-3}{*}{Gender} &
  LP &
  \cellcolor[HTML]{CBCEFB}1.916451 &
  1.127964 &
  0.938095 &
  1.19075 &
  \cellcolor[HTML]{CBCEFB}2.194655 \\ \cline{2-8} 
 &
   &
  SSL &
  1.151525 &
  0.949581 &
  \cellcolor[HTML]{CBCEFB}1.313889 &
  \cellcolor[HTML]{CBCEFB}2.009774 &
  1.143114 \\ \cline{3-8} 
 &
   &
  Sup. &
  \cellcolor[HTML]{CBCEFB}1.221075 &
  0.956821 &
  1.173115 &
  \cellcolor[HTML]{CBCEFB}1.788963 &
  1.221397 \\ \cline{3-8} 
 &
  \multirow{-3}{*}{Insurance} &
  LP &
  1.055201 &
  0.944077 &
  1.052997 &
  \cellcolor[HTML]{CBCEFB}1.542647 &
  1.041421 \\ \cline{2-8} 
 &
   &
  SSL &
  \cellcolor[HTML]{CBCEFB}1.387281 &
  0.873203 &
  0.952534 &
  \cellcolor[HTML]{CBCEFB}2.064516 &
  \cellcolor[HTML]{CBCEFB}1.318247 \\ \cline{3-8} 
 &
   &
  Sup. &
  \cellcolor[HTML]{CBCEFB}1.410703 &
  0.85925 &
  1.018682 &
  \cellcolor[HTML]{CBCEFB}2.124872 &
  \cellcolor[HTML]{CBCEFB}1.319083 \\ \cline{3-8} 
\multirow{-15}{*}{MIMIC} &
  \multirow{-3}{*}{Language} &
  LP &
  \cellcolor[HTML]{CBCEFB}0.646858 &
  0.807558 &
  \cellcolor[HTML]{CBCEFB}1.222419 &
  \cellcolor[HTML]{CBCEFB}2.149763 &
  \cellcolor[HTML]{CBCEFB}0.568459 \\ \hline
 &
   &
  SSL &
  0.998046 &
  0.999164 &
  0.992309 &
  0.948194 &
  0.983189 \\ \cline{3-8} 
 &
   &
  Sup. &
  0.993162 &
  1.000327 &
  0.995877 &
  0.944088 &
  0.979517 \\ \cline{3-8} 
 &
  \multirow{-3}{*}{Age} &
  LP &
  1.009543 &
  0.997778 &
  0.978821 &
  0.954434 &
  0.993137 \\ \cline{2-8} 
 &
   &
  SSL &
  0.990607 &
  0.997288 &
  0.992428 &
  0.965692 &
  0.98379 \\ \cline{3-8} 
 &
   &
  Sup. &
  0.992224 &
  0.998345 &
  0.99491 &
  0.970256 &
  0.98644 \\ \cline{3-8} 
 &
  \multirow{-3}{*}{Ethnicity} &
  LP &
  1.017328 &
  0.990076 &
  0.962152 &
  0.98008 &
  1.00302 \\ \cline{2-8} 
 &
   &
  SSL &
  1.081529 &
  1.014118 &
  1.007385 &
  0.991359 &
  1.047817 \\ \cline{3-8} 
 &
   &
  Sup. &
  1.071921 &
  1.015271 &
  1.011415 &
  0.984422 &
  1.039689 \\ \cline{3-8} 
\multirow{-9}{*}{MESA} &
  \multirow{-3}{*}{Gender} &
  LP &
  1.033791 &
  1.021893 &
  1.017314 &
  0.941302 &
  1.009246 \\ \hline
 &
   &
  SSL &
  0.935372 &
  \cellcolor[HTML]{CBCEFB}1.348379 &
  \cellcolor[HTML]{CBCEFB}1.479401 &
  1.015322 &
  1.084999 \\ \cline{3-8} 
 &
   &
  Sup. &
  \cellcolor[HTML]{CBCEFB}0.767811 &
  1.007267 &
  0.95688 &
  \cellcolor[HTML]{CBCEFB}0.55 &
  \cellcolor[HTML]{CBCEFB}0.665323 \\ \cline{3-8} 
 &
  \multirow{-3}{*}{Disability} &
  LP &
  \cellcolor[HTML]{CBCEFB}0.635082 &
  \cellcolor[HTML]{CBCEFB}1.27369 &
  \cellcolor[HTML]{CBCEFB}1.388889 &
  \cellcolor[HTML]{CBCEFB}0.793521 &
  \cellcolor[HTML]{CBCEFB}0.695868 \\ \cline{2-8} 
 &
   &
  SSL &
  1.177066 &
  0.911646 &
  0.856541 &
  1.200049 &
  \cellcolor[HTML]{CBCEFB}1.183419 \\ \cline{3-8} 
 &
   &
  Sup. &
  1.14846 &
  1.057243 &
  1.188105 &
  \cellcolor[HTML]{CBCEFB}1.650568 &
  \cellcolor[HTML]{CBCEFB}1.339067 \\ \cline{3-8} 
 &
  \multirow{-3}{*}{Ethnicity} &
  LP &
  0.877829 &
  0.892041 &
  1.042793 &
  1.118649 &
  0.863588 \\ \cline{2-8} 
 &
   &
  SSL &
  0.993029 &
  \cellcolor[HTML]{CBCEFB}0.780661 &
  1.07721 &
  \cellcolor[HTML]{CBCEFB}1.694643 &
  1.005157 \\ \cline{3-8} 
 &
   &
  Sup. &
  1.006579 &
  \cellcolor[HTML]{CBCEFB}0.786325 &
  1.085652 &
  \cellcolor[HTML]{CBCEFB}1.734266 &
  1.02626 \\ \cline{3-8} 
\multirow{-9}{*}{GLOBEM} &
  \multirow{-3}{*}{Gender} &
  LP &
  1.159948 &
  0.850333 &
  0.986795 &
  \cellcolor[HTML]{CBCEFB}1.747962 &
  \cellcolor[HTML]{CBCEFB}1.278902 \\ \hline
\end{tabular}%
}
\end{table}
\rev{Beyond the provided meta-metric, Table~\ref{tab:fairness-metrics} presents the values of individual fairness metrics, where we compare the best-performing SSL model, with the linear probing model (i.e., the model where we freeze all layers of the pre-trained SSL model and only add a small classification head on top to predict the target labels) as an SSL baseline, and the supervised model (Sup). Partially freezing some layers while fine-tuning others allows for preserving the debiased pre-trained representations to an extent, while allowing specialization of some layers to the target distribution to maintain competitive accuracy. We see that the SSL model (with some level of judicious fine-tuning) shows superior performance for the MIMIC and GLOBEM datasets, having the maximum or equal number of within-range values for the studied fairness metrics for all protected attributes.}

\end{document}